\documentclass[9pt,a4paper,twocolumn]{extarticle}

\usepackage[margin=0.75in,columnsep=0.25in]{geometry}
\usepackage{times}
\usepackage{microtype}
\usepackage[numbers,sort&compress]{natbib}
\usepackage{booktabs}
\usepackage{multirow}
\usepackage{tabularx}
\usepackage{amsmath}
\usepackage{amsfonts}
\usepackage{amssymb}
\usepackage{placeins}
\usepackage{xurl}
\usepackage[colorlinks=true,linkcolor=blue,citecolor=blue,urlcolor=blue]{hyperref}
\usepackage{float}
\usepackage{caption}
\usepackage{graphicx}
\usepackage{xcolor}
\usepackage{eurosym}
\usepackage{authblk}
\usepackage{tikz}
\usepackage{pgfplots}
\pgfplotsset{compat=1.18}
\usetikzlibrary{shapes.geometric, arrows.meta, positioning, calc}

\definecolor{gpu5060}{RGB}{66,133,244}
\definecolor{gpu5070}{RGB}{52,168,83}
\definecolor{gpu5090}{RGB}{234,67,53}
\definecolor{gpu5090x2}{RGB}{251,188,5}

\title{\textbf{Private LLM Inference on Consumer Blackwell GPUs:\\ A Practical Guide for Cost-Effective Local Deployment in SMEs}}

\author[1]{Jonathan Knoop}
\author[2]{Hendrik Holtmann}
\affil[1]{IE Business University, Madrid, Spain (jmknoop.ieu2025@student.ie.edu)}
\affil[2]{Independent Researcher, Hamburg, Germany (holtmann@gmail.com)}

\date{}

\begin{document}

\twocolumn[
  \maketitle

\begin{abstract}
SMEs increasingly seek alternatives to cloud LLM APIs, which raise data privacy concerns. Dedicated cloud GPU instances offer improved privacy but with limited guarantees and ongoing costs, while professional on-premise hardware (A100, H100) remains prohibitively expensive.
We present a systematic evaluation of NVIDIA's Blackwell consumer GPUs (RTX 5060~Ti, 5070~Ti, 5090) for production LLM inference, benchmarking four open-weight models (Qwen3-8B, Gemma3-12B, Gemma3-27B, GPT-OSS-20B) across 79 configurations spanning quantization formats (BF16, W4A16, NVFP4, MXFP4), context lengths (8k--64k), and three workloads: RAG, multi-LoRA agentic serving, and high-concurrency APIs.
The RTX 5090 delivers 3.5--4.6$\times$ higher throughput than the 5060~Ti with 21$\times$ lower latency for RAG, but budget GPUs achieve the highest throughput-per-dollar for API workloads with sub-second latency.
NVFP4 quantization provides 1.6$\times$ throughput over BF16 with 41\% energy reduction and only 2--4\% quality loss.
Self-hosted inference costs \$0.001--0.04 per million tokens (electricity only)---40--200$\times$ cheaper than budget-tier cloud APIs---with hardware breaking even in under four months at moderate volume (30M tokens/day).
Our results show that consumer GPUs can reliably replace cloud inference for most SME workloads, except latency-critical long-context RAG, where high-end GPUs remain essential.
We provide deployment guidance and release all benchmark data for reproducible SME-scale deployments.
\end{abstract}

  \vspace{1.5em}
  \section{Introduction}
  \label{sec:intro}
]


Large language models (LLMs) have become indispensable productivity tools, with frontier models from OpenAI~\cite{openai2023gpt4}, Google~\cite{google2024gemini}, and Anthropic~\cite{anthropic2024claude} serving hundreds of millions of users.
Small and medium-sized enterprises (SMEs) increasingly deploy LLM-powered solutions for customer service, knowledge management, and document processing (AI adoption among EU enterprises grew 67\% in 2024~\cite{eurostat2025ai}).
However, two barriers impede deployment: data privacy concerns with third-party APIs, and the high cost or limited availability of professional GPU infrastructure.

Commercial APIs offer quality inference with minimal overhead, but introduce compliance risks for enterprises handling sensitive data---third-party inference raises concerns about data leakage and regulatory compliance (GDPR, HIPAA)~\cite{bommasani2022foundation}.
Local deployment addresses these concerns by keeping data on-premises with predictable costs, yet building production-grade systems requires expertise in hardware selection, model optimization, and serving infrastructure.

NVIDIA's Blackwell consumer GPUs (RTX 50-series, early 2025) represent a potential inflection point for democratizing LLM deployment.
These cards offer improved memory bandwidth (up to 1.8~TB/s on RTX~5090), native 4-bit inference (NVFP4), and favorable pricing (\$2,000 vs.\ \$25,000+ for H100).
Unlike datacenter GPUs, consumer cards are readily available through retail and require no specialized infrastructure.
However, existing benchmarks focus on training or datacenter accelerators~\cite{mattson2020mlperf}, leaving a gap in understanding consumer GPU performance for production inference.

This paper addresses this gap through empirical study of LLM inference on NVIDIA's RTX 5060~Ti, 5070~Ti, and 5090.
We benchmark four open-weight models (Qwen3-8B, Gemma3-12B, Gemma3-27B, GPT-OSS-20B) across 79 configurations spanning quantization formats (BF16, W4A16, NVFP4, MXFP4), context lengths (8k--64k), and concurrency levels (4--256 requests).
Using vLLM~\cite{kwon2023vllm} and AIPerf~\cite{nvidia2024aiperf}, we measure throughput, latency, and energy consumption across three workloads: RAG, multi-LoRA agentic serving, and high-concurrency APIs.
Our analysis reveals self-hosted inference achieves cost parity with commercial APIs within 1--4 months at moderate usage (30M tokens/day), with subsequent operation at 40--200$\times$ lower cost than budget-tier cloud models.
All code, configurations, and Docker images are released for reproducibility.

Section~\ref{sec:related} reviews related work; Section~\ref{sec:methods} presents methodology; Section~\ref{sec:results} reports results and practical recommendations; Section~\ref{sec:limitations} discusses limitations; Section~\ref{sec:conclusion} summarizes key takeaways.

\bigskip
\section{Related Work}
\label{sec:related}

\paragraph{LLM Inference Optimization.}
Efficient LLM serving has attracted significant research attention. vLLM~\cite{kwon2023vllm} introduced PagedAttention for memory-efficient KV-cache management, enabling higher throughput through continuous batching. Quantization techniques including AWQ~\cite{lin2023awq}, QLoRA~\cite{dettmers2023qlora}, and hardware-specific formats like NVFP4~\cite{nvidia2025nvfp4} reduce memory footprint while preserving model quality. KV-cache quantization~\cite{hooper2024kvquant} further extends context length capabilities on memory-constrained devices. Our work builds on these advances by empirically characterizing their combined effectiveness on consumer hardware.

\paragraph{GPU Benchmarking for Machine Learning.}
MLPerf~\cite{mattson2020mlperf} established standardized benchmarks for ML training and inference on datacenter hardware, but focuses primarily on professional GPUs (A100, H100) rather than consumer cards. NVIDIA's AIPerf~\cite{nvidia2024aiperf} provides inference benchmarking tools but published results target datacenter deployments. Prior consumer GPU evaluations have focused on gaming or cryptocurrency workloads rather than production LLM inference. Our study addresses this gap by providing a systematic, SME-focused characterization of Blackwell-generation consumer GPUs across realistic production workloads including multi-LoRA serving and quality-aware deployment decisions.

\paragraph{Cost-Effective LLM Deployment.}
The tension between cloud API convenience and local deployment economics has driven interest in efficient self-hosting. Foundation model analyses~\cite{bommasani2022foundation} highlight data governance concerns with third-party APIs. While cloud providers offer serverless inference, the marginal cost structure disadvantages high-volume users. Our break-even analysis quantifies when self-hosted inference becomes economically advantageous, complementing qualitative discussions of deployment trade-offs in the literature.

\section{Experimental Methodology}
\label{sec:methods}

We evaluate whether consumer GPUs---NVIDIA's RTX~5060~Ti~16\,GB, RTX~5070~Ti~16\,GB, and RTX~5090~32\,GB---can serve modern LLMs for typical SME workloads.
We focus on three deployment scenarios:
(i) retrieval-augmented generation (RAG) with long contexts,
(ii) multi-LoRA agentic workloads with frequent adapter switching, and
(iii) high-concurrency API serving.
Experiments use vLLM~\cite{kwon2023vllm} and AIPerf.

\subsection{Research Questions}

Our methodology is structured around the following research questions (RQs):

\begin{itemize}
  \item \textbf{RQ1 (Throughput \& Latency).} For the three workload classes (RAG, agentic multi-LoRA, and high-concurrency API use), how do consumer Blackwell GPUs (RTX~5060~Ti, 5070~Ti, 5090) compare in terms of tokens-per-second (TPS), time-to-first-token (TTFT), and tail latencies across single- and dual-GPU configurations?
  \item \textbf{RQ2 (Quantization Trade-offs).} How do different low-precision formats---4-bit weight-only (W4A16 via AWQ~\cite{lin2023awq}), mixed-precision NVFP4~\cite{nvidia2025nvfp4}, and MXFP4~\cite{microsoft2024mxfp4}---affect throughput, memory footprint, and energy per generated token, relative to higher-precision baselines where these fit in memory?
  \item \textbf{RQ3 (Agentic Overheads).} What is the overhead of frequent LoRA adapter switching---a proxy for multi-agent systems and tenant-specific models---and how well does vLLM's adapter management scale on commodity hardware?
  \item \textbf{RQ4 (Energy \& Cost).} For SME-relevant local deployments, what is the energy consumption (Wh/MTok) and estimated electricity cost per million tokens across different GPU configurations and quantization schemes? Which combinations offer the best energy efficiency for each workload class?
\end{itemize}

\subsection{Model Suite}

We benchmark four recent open(-weight) LLMs from three model families that explicitly target efficient deployment on commodity accelerators and are suitable for SME-scale workloads.
To ensure representation across the global open-weight ecosystem, we include models from Chinese (Qwen3) and US (Gemma3, GPT-OSS) organizations:

\begin{itemize}
  \item \textbf{Qwen3-8B.} Qwen3-8B is a dense 8.2B-parameter model from the Qwen3 family~\cite{yang2025qwen3}, supporting context lengths up to 128k tokens via YaRN~\cite{peng2023yarn} and offering strong multilingual and instruction-following performance. We use the instruction-tuned variant with thinking mode disabled as our primary candidate for adapter-based customization in the agentic scenario (Section~\ref{subsec:agentic-workload}).
  \item \textbf{Gemma3-12B and 27B.} Google's Gemma3 models are designed to run efficiently on single GPUs while providing strong general knowledge and language understanding~\cite{gemma32025}. We use the 12B and 27B instruction-tuned variants and their official tokenizer and chat templates.
  \item \textbf{GPT-OSS-20B (reasoning MoE).} OpenAI's GPT-OSS-20B is an open-weight mixture-of-experts (MoE) model optimized for reasoning on consumer hardware with $\geq$16\,GB memory~\cite{openai2025gptoss,openai2025gptossmodelcard,news2025gptoss_tomshardware}. Its sparse architecture activates only a subset of experts per token, reducing effective compute cost below its 20B parameter count. This makes it attractive for SME deployments requiring stronger reasoning than comparable sized dense models.
\end{itemize}

These models span compact dense (Qwen3-8B), medium dense (Gemma3-12B), large dense (Gemma3-27B), and reasoning-focused MoE (GPT-OSS-20B) architectures. This diversity lets us examine how model architecture interacts with quantization, context length, and GPU memory.

\subsection{Quantization Schemes}
\label{subsec:quantization}

Uncompressed BF16/FP16 or FP8 variants of these models typically require well beyond 16\,GB of device memory at practical context lengths (e.g., $\geq$16k tokens), and even 32\,GB can be constraining when serving multiple concurrent requests.
To make deployment realistic on consumer GPUs, we rely on post-training quantization and mixed-precision formats.

We evaluate three quantization formats, all of which are supported by vLLM and compatible with our target hardware:

\begin{itemize}
  \item \textbf{W4A16 (via AWQ).} 4-bit weight-only quantization with 16-bit activations, using Activation-aware Weight Quantization~\cite{lin2023awq}. AWQ rescales salient weight channels to reduce quantization error and has become the de-facto standard for local inference. We use RedHat's W4A16 checkpoints for Gemma3.\footnote{Available at \url{https://huggingface.co/RedHatAI}; e.g., \texttt{gemma-3-27b-it-quantized.w4a16}.}
  \item \textbf{NVFP4 (weights and activations).} NVIDIA's 4-bit floating-point format with native hardware acceleration on Blackwell Tensor Cores~\cite{nvidia2025nvfp4}, combining E2M1 representation with dual-level scaling (FP8 micro-blocks, FP32 per-tensor). NVFP4 quantizes both weights and activations, reportedly matching FP8 accuracy while achieving 2.3$\times$ higher throughput than weight-only 4-bit methods like AWQ~\cite{marie2025nvfp4,lowprecision2025nvfp4}. We evaluate NVFP4 for Qwen3-8B, Gemma3-12B, and Gemma3-27B.\footnote{For Qwen3-8B we use NVIDIA's official NVFP4 checkpoint (\url{https://huggingface.co/nvidia/Qwen3-8B-NVFP4}). For Gemma3, we quantize using llm-compressor (\url{https://github.com/vllm-project/llm-compressor}); scripts are provided in our code release.}
  \item \textbf{MXFP4 (microscaling).} Microsoft's portable 4-bit microscaling format~\cite{microsoft2024mxfp4} using per-block scaling. Unlike Blackwell-specific NVFP4, MXFP4 works across hardware platforms. We use MXFP4 for GPT-OSS-20B following OpenAI's recommendation~\cite{openai2025gptoss,openai2025gptossmodelcard}, targeting 16\,GB-class consumer GPUs~\cite{news2025gptoss_tomshardware}.
\end{itemize}

For each model and GPU configuration, we select the most aggressive quantization that (i) fits within available device memory for the targeted context length, and (ii) does not degrade task accuracy by more than 2--4 percentage points (cf.\ Section~\ref{subsec:task-eval}).
Where memory permits (Qwen3-8B and Gemma3-12B on RTX~5090), we run BF16 baselines to quantify quality and latency impacts.

\paragraph{KV Cache Precision.}
We use vLLM's default FP16 KV cache in most configurations, enabling FP8 KV cache quantization only where memory constraints would otherwise prevent deployment.
Evaluations show FP8 quantization achieves $>$99\% accuracy preservation on standard benchmarks (MMLU, HellaSwag, GSM8k)~\cite{neuralmagic2024fp8}, making it a practical choice for extending context length on memory-constrained hardware.
For example, FP8 KV cache enables 32k--64k context deployment on 16\,GB consumer GPUs that would otherwise be limited to shorter contexts.
FP8 KV cache is used for all dual-GPU configurations with long context ($\geq$16k) or high-concurrency agentic workloads, as well as GPT-OSS-20B deployments on budget hardware.

\subsection{Hardware Platforms and Hosting}

We run all experiments on bare-metal consumer GPUs rented through VAST.ai, a peer-to-peer GPU rental marketplace.
We use three GPU SKUs, each in a single- and dual-GPU configuration:

\begin{itemize}
  \item \textbf{RTX~5060~Ti 16\,GB.} Lower-midrange Blackwell GPU with 16\,GB GDDR7 and fifth-gen Tensor Cores, typical of SME workstations.
  \item \textbf{RTX~5070~Ti 16\,GB.} Upper-midrange Blackwell with 16\,GB GDDR7, higher compute and memory bandwidth than 5060~Ti.
  \item \textbf{RTX~5090 32\,GB.} Flagship Blackwell with 32\,GB GDDR7, representing high-end SME workstation budgets~\cite{nvidia5090product}.
\end{itemize}

For each SKU, we select instances whose CPU, RAM, and storage characteristics are sufficient to avoid obvious bottlenecks (e.g., at least 8 CPU cores and 64\,GB host RAM for the dual-GPU setups).
We configure all instances with NVIDIA driver version 570.x or later, CUDA~12.9, and cuDNN~9.x, which are required for vLLM's NVFP4 kernel support on Blackwell GPUs.
Where possible, both single- and dual-GPU configurations run on the same host model to reduce confounding factors such as PCIe bandwidth.

\paragraph{Dual-GPU Parallelism Strategy.}
For dual-GPU configurations, we use \emph{tensor parallelism} over pipeline parallelism.
Tensor parallelism partitions weight matrices and activations across GPUs with all-reduce communication per layer~\cite{shoeybi2019megatron}.
We prefer this because: (i)~pipeline parallelism is poorly suited to autoregressive decoding where generated tokens must cycle back through stages; (ii)~tensor parallelism achieves lower per-request latency via full parallelization; and (iii)~vLLM recommends tensor parallelism for single-node multi-GPU configurations~\cite{kwon2023vllm}.
We use \texttt{-{}-tensor-parallel-size=2} for all dual-GPU experiments.

\subsection{Software Stack}

\paragraph{Inference Engine.}
All models are served using vLLM~0.12~\cite{kwon2023vllm}, a production-grade LLM inference engine that implements PagedAttention for efficient KV-cache management and supports a wide range of quantization formats and multi-LoRA adapters.
We selected vLLM over alternatives (TensorRT-LLM, SGLang, llama.cpp) for its combination of production maturity, early Blackwell/NVFP4 support, and native multi-LoRA serving---features essential for our agentic workload evaluation.
vLLM is compiled with CUDA~12.9 support for Blackwell GPUs and NVFP4 kernels.
We use vLLM's default dynamic batching configuration, which automatically adjusts the number of concurrent sequences and batched tokens based on available GPU memory and KV-cache capacity.
We use vLLM's default for \texttt{max\_num\_seqs} (maximum concurrent sequences), reducing it only when startup fails with out-of-memory errors.
For \texttt{max\_num\_batched\_tokens}, we set a fixed value of 8192 across all experiments (4$\times$ the vLLM default of 2048) to improve prefill performance for long-context workloads, trading memory headroom for faster time-to-first-token.
This configuration reflects a practical optimization for production deployments prioritizing responsiveness.

\paragraph{Benchmark Harness.}
We use AIPerf~0.3.0~\cite{nvidia2024aiperf} as the primary benchmark harness.
AIPerf stress-tests generative model serving stacks with configurable request distributions, concurrency levels, and detailed client-side metrics such as latency and per-token throughput.
AIPerf natively supports integration with vLLM and other inference servers via HTTP and gRPC.

To simulate agent-like workloads that switch frequently between LoRA adapters, we use the AIPerf model-selection strategy. This option randomly assigns one of the fine-tuned models to each request, effectively mimicking rapid adapter changes under real-world conditions.

\paragraph{GPU Telemetry and Energy Measurement.}

We collect GPU metrics via AIPerf's integration with NVIDIA DCGM~\cite{nvidia_dcgm_docs} and DCGM-Exporter~\cite{nvidia_dcgm_exporter,nvidia_dcgm_exporter_docs}, which exposes power draw, SM utilization, memory bandwidth, and temperature through a Prometheus-compatible endpoint that AIPerf scrapes during benchmark runs.\footnote{Exact telemetry configuration will be released with benchmark code.}

We compute energy consumption per run as:
\begin{equation}
  E = \sum_{i=1}^{N} P(t_i) \cdot \Delta t \quad [\mathrm{Joules}],
\end{equation}
where $P(t_i)$ is power at sample $i$, $\Delta t$ is the sampling interval, and $N$ is the total number of samples.
For dual-GPU configs, we sum power draw from both GPUs.
Given output tokens $N_{\text{tok}}$, we derive Wh/MTok as our primary energy metric.

For electricity cost, we compute:
\begin{equation}
  \mathrm{Cost}_{\text{energy}} = \frac{E_{\text{Wh/MTok}} \times r}{1000} \quad [\$/\mathrm{MTok}].
\end{equation}
We use $r = \$0.12$/kWh (US commercial average) as our baseline rate. For European deployments, electricity costs are typically higher: the EU-27 average for non-household consumers is approximately \euro0.25/kWh (\$0.27/kWh), with significant variation from \euro0.10/kWh in Scandinavia to \euro0.40/kWh in Germany and Italy.
To account for this range, practitioners should scale our reported costs proportionally for their local electricity rates.
All cost figures in this paper use the US baseline; European operators should scale costs by approximately 2--2.5$\times$ for Western European rates.
This captures marginal operational cost only. For total cost of ownership, hardware amortization adds \$0.01--0.05/MTok over a 2-year lifespan depending on GPU tier and utilization, bringing all-in self-hosted cost to approximately \$0.02--0.09/MTok.

\subsection{Workload Scenarios}

We define three synthetic but SME-realistic workloads, each aligned with a typical deployment pattern.

\subsubsection{RAG with Long Contexts}
\label{subsec:rag-workload}

The RAG scenario approximates internal question answering over domain-specific documents (e.g., contracts, knowledge bases, product manuals).
We assume that document retrieval and chunking have already been performed upstream, and that the LLM is invoked with a concatenation of retrieved passages plus the user question.

For each model, we configure AIPerf to generate requests with the following properties:

\begin{itemize}
  \item \emph{Prompt length:} We evaluate four synthetic context length tiers to represent typical to very long RAG-style inputs:
  \begin{itemize}
    \item 8{,}192 tokens (short-RAG)
    \item 16{,}384 tokens (medium-RAG)
    \item 32{,}768 tokens (long-RAG)
    \item 65{,}536 tokens (very-long-RAG)
  \end{itemize}
  Each tier is generated using AIPerf’s synthetic input mode. Not all models and GPU configurations can support all lengths at practical concurrency levels; feasible limits are reported in Section~\ref{subsec:context-feasibility}.
  
  \item \emph{Maximum output length:} 512 tokens, enforced via AIPerf’s output-token constraints.
  
  \item \emph{Inference parameters:} Temperature and sampling settings follow model-card recommendations for instruction-following models (e.g., temperature=0.7, top-p=0.9).
  
  \item \emph{Concurrency:} Fixed maximum concurrency levels of 4, 8, and 16 concurrent requests, representing small-team and light multi-tenant usage scenarios.
  
  \item \emph{Request arrival pattern:} Poisson arrivals using AIPerf’s request-rate mode with a maximum concurrency cap. The request rate is tuned so that the GPU reaches 80--90\% steady-state utilization at the highest concurrency level, avoiding queue overflow while still stressing the system.
\end{itemize}

\subsubsection{Agentic Multi-LoRA Workloads}
\label{subsec:agentic-workload}

The agentic scenario approximates SMEs that deploy multiple fine-tuned adapters for distinct tasks or departments (e.g., customer support, legal drafting, code assistance), all sharing a common base model.
This pattern is common in LoRA-based fine-tuning frameworks such as QLoRA~\cite{dettmers2023qlora}, where many low-rank adapters can be hosted simultaneously on a single GPU.

We construct this workload as follows:

\begin{itemize}
  \item \emph{Base model:} Qwen3-8B, whose compact size maximizes VRAM for hosting multiple adapters on 16\,GB GPUs.
  \item \emph{Adapters:} a pool of three LoRA adapters, each fine-tuned offline on separate datasets: (i) customer support FAQ, (ii) technical document drafting, and (iii) structured JSON output generation.
  \item \emph{Adapter switching:} each request is assigned an adapter uniformly at random from the pool, simulating multi-tenant traffic where different departments or tools hit the same endpoint.
  \item \emph{Prompt length:} 2{,}048 tokens average (task instructions plus short context), with a maximum of 4{,}096 tokens.
  \item \emph{Maximum output length:} 512 tokens.
  \item \emph{Concurrency:} we target higher concurrency than in the RAG setting, evaluating 16, 32, and 64 concurrent requests per GPU configuration.
\end{itemize}

We report both aggregate metrics and adapter-specific breakdowns.
In particular, we examine whether frequent adapter switching leads to cache thrashing, higher TTFT, or reduced throughput on single- versus dual-GPU setups, and how this interacts with different quantization formats (W4A16, NVFP4, and MXFP4).

\subsubsection{High-Concurrency API Serving}
\label{subsec:api-workload}

The API workload corresponds to short-prompt, latency-sensitive use cases such as chatbots, lightweight classification, or auto-completion that SMEs might embed into internal tools or public-facing websites.
In this setting, concurrency rather than context length is the primary bottleneck.

We configure AIPerf as follows:

\begin{itemize}
  \item \emph{Prompt length:} 128--512 tokens (median 256), representing short chat turns or classification inputs.
  \item \emph{Maximum output length:} 256 tokens.
  \item \emph{Concurrency:} scalable concurrency sweeps from 32 up to 256 concurrent requests, increasing until the system becomes unstable (e.g., excessive queueing or timeouts).
  \item \emph{Request arrival pattern:} open-loop Poisson arrivals with a configurable mean rate to explore low-, medium-, and high-load regimes.
\end{itemize}

In this scenario we deploy all four models, but expect smaller models and aggressive quantization (e.g., NVFP4 or MXFP4) to yield the best cost-per-token under strict latency constraints.

\subsection{Context Length Feasibility Analysis}
\label{subsec:context-feasibility}

While models advertise support for very long contexts (e.g. Gemma3 up to 128K), actual usable context length depends on GPU memory, KV-cache size, and concurrency requirements.
We determine the \emph{maximum feasible context} for each (model, precision, GPU) configuration via stress testing: incrementally increasing context length (8k → 16k → 32k → 64k → 128k) at fixed concurrency of 4 requests, recording the longest context that completes without OOM or timeouts.

Maximum context must satisfy: (i) model weights + KV-cache for $\geq$4 concurrent requests fit in VRAM, and (ii) no excessive queueing or failures.
Results are reported in the RAG throughput table (Table~\ref{tab:rag-throughput}).
For the RAG workload (Section~\ref{subsec:rag-workload}), we only evaluate contexts up to the feasible limit per configuration.

\subsection{Task-aligned Quality Evaluation}
\label{subsec:task-eval}

To ensure that our recommended configurations do not suffer unacceptable quality degradation, we complement system-level metrics with task-aligned quality benchmarks. Importantly, the goal is not to compare models against each other---such comparisons are confounded by differences in training data, architecture, and scale---but rather to quantify the accuracy loss introduced by quantization relative to each model's own higher-precision baseline.

Quality is evaluated per (model, precision) configuration on a reference GPU where the model fits comfortably. Our evaluation settings (few-shot counts, prompt templates, sampling parameters) may differ from model authors' published results; practitioners should focus on the delta between BF16 and quantized variants rather than absolute scores.

We use widely-recognized benchmarks aligned with each workload scenario:

\begin{itemize}
  \item \textbf{RAG scenario.} We evaluate on \textbf{MMLU}~\cite{hendrycks2021mmlu} (5-shot, 500 examples), the de-facto standard for measuring knowledge and reasoning across 57 subjects, directly relevant to enterprise QA.

  \item \textbf{Agentic scenario.} We use \textbf{GSM8K}~\cite{cobbe2021gsm8k} (full test set, 1,319 problems, chain-of-thought), a standard benchmark for multi-step math reasoning that tests problem decomposition and sequential inference capabilities critical for agent-based systems.

  \item \textbf{API scenario.} We evaluate on \textbf{HellaSwag}~\cite{zellers2019hellaswag} (0-shot, 10,042 examples), a widely-used commonsense reasoning benchmark representative of diverse short-form tasks typical in API endpoints.
\end{itemize}

All evaluations are conducted using the Language Model Evaluation Harness (lm-eval)~\cite{eval-harness} v0.4.9, which provides a unified framework for reproducible few-shot evaluation of language models.

For each benchmark, we compare quantized models (W4A16, NVFP4) against BF16 baselines where memory permits to quantify quality trade-offs. We note two exceptions: (1)~GPT-OSS-20B is excluded from quality evaluation because it is released exclusively in MXFP4 as its native format---no BF16 or other precision weights are available for comparison~\cite{openai2025gptoss,openai2025gptossmodelcard}. Additionally, lm-eval does not fully support OpenAI's Harmony chat format; we retain GPT-OSS-20B in throughput and latency benchmarks where it provides valid performance data. (2)~Gemma3-27B BF16 baseline exceeds available GPU memory in all tested configurations; we report only quantized results for this model, comparing W4A16 against NVFP4.
Results are reported in Tables~\ref{tab:rag-quality}--\ref{tab:api-quality}.

\subsection{Performance Metrics}

Across all scenarios we collect the following system-level metrics:

\begin{itemize}
  \item \textbf{Throughput.} Total output tokens per second (global TPS) and per-request TPS as reported by AIPerf.
  \item \textbf{Latency.} Time-to-first-token (TTFT), median latency, and tail latencies (P95 and P99) for each scenario and concurrency level.
  \item \textbf{Energy.} Joules per output token and Wh per million output tokens (Wh/MTok), derived from DCGM power telemetry as described above.
\end{itemize}

\paragraph{Interpreting Throughput for Capacity Planning.}
To estimate per-user experience from aggregate throughput, divide TPS by concurrency. For example, 488~TPS at 64 concurrent users yields approximately 7.6~tokens/user/second. We report TPS/User explicitly in multi-tenant scenarios (Table~\ref{tab:rq3-agentic}); for other tables, readers can derive per-user rates using this formula.

Tables~\ref{tab:rag-throughput}--\ref{tab:energy-efficiency} summarize system-level results; detailed per-run logs and plots will be released as supplementary material. Not all models fit on all GPUs; shown are only the feasible model-GPU combinations.

\paragraph{Reproducibility and Extensibility.}
To enable full reproducibility and allow practitioners to benchmark new models or hardware configurations, we release:
\begin{itemize}
    \item A \textbf{Docker image} containing the complete benchmark environment (vLLM, AIPerf, DCGM telemetry, and all dependencies) pre-configured for Blackwell GPUs.
    \item A \textbf{GitHub repository} with benchmark orchestration scripts, configuration files for all 79 tested configurations, and instructions for extending benchmarks to new models or GPU SKUs.
    \item \textbf{Raw benchmark data} including per-run JSON logs, energy traces, and quality evaluation outputs.
\end{itemize}
The Docker image enables one-command benchmark execution on any compatible GPU, while the repository allows researchers to rebuild or customize the environment for their specific requirements.\footnote{Docker image: \url{https://hub.docker.com/r/holtmann/llm-benchmark}. GitHub repository: \url{https://github.com/hholtmann/llm-consumer-gpu-benchmark}.}

\section{Results and Discussion}
\label{sec:results}

We report task-aligned quality evaluation results to verify that quantization does not cause unacceptable accuracy loss, followed by system-level performance metrics (throughput, latency, energy).
\subsection{Task-Aligned Quality Evaluation}

\begin{table}[H]
  \centering\small
  \caption{RAG-style quality on MMLU (5-shot, 500 examples).
  Accuracy is reported as fractions (0--1 scale) with 95\% confidence intervals.
  Models are evaluated once per (model, precision) configuration on a reference GPU.
  ``--'' indicates context length not supported or benchmark not applicable.}
  \label{tab:rag-quality}
  \begin{tabular}{l l l c}
    \toprule
    Model & Params (B) & Precision & MMLU (5-shot) \\
    \midrule
    Qwen3-8B & 8.2 & NVFP4 & 0.7509 ± 0.0038 \\
    Qwen3-8B & 8.2 & BF16 (baseline) & 0.7729 ± 0.0036 \\
    Gemma3-12B & 12 & W4A16 & 0.5904 ± 0.0043 \\
    Gemma3-12B & 12 & NVFP4 & 0.5795 ± 0.0043 \\
    Gemma3-12B & 12 & BF16 (baseline) & 0.6202 ± 0.0042 \\
    Gemma3-27B$^\ddagger$ & 27 & W4A16 & 0.6910 ± 0.0040 \\
    Gemma3-27B$^\ddagger$ & 27 & NVFP4 & 0.6816 ± 0.0040 \\
    \bottomrule
    \multicolumn{4}{l}{\footnotesize $^\ddagger$BF16 baseline exceeds GPU memory; quantized results only.}
  \end{tabular}
\end{table}
\vspace{-10pt}

\begin{table}[H]
  \centering\small
  \caption{Agentic reasoning quality on GSM8K (1,319 problems, chain-of-thought). Accuracy is the fraction of problems with exact numeric match to reference.}
  \label{tab:agent-quality}
  \begin{tabular}{l l l c}
    \toprule
    Model & Params (B) & Precision & GSM8K Accuracy \\
    \midrule
    Qwen3-8B & 8.2 & NVFP4 & 0.8711 ± 0.0092 \\
    Qwen3-8B & 8.2 & BF16 (baseline) & 0.8901 ± 0.0086 \\
    Gemma3-12B & 12 & W4A16 & 0.8203 ± 0.0106 \\
    Gemma3-12B & 12 & NVFP4 & 0.7892 ± 0.0112  \\
    Gemma3-12B & 12 & BF16 (baseline) & 0.8249 ± 0.0105 \\
    Gemma3-27B$^\ddagger$ & 27 & W4A16 & 0.8886 ± 0.0087  \\
    Gemma3-27B$^\ddagger$ & 27 & NVFP4 & 0.8635 ± 0.0095 \\
    \bottomrule
    \multicolumn{4}{l}{\footnotesize $^\ddagger$BF16 baseline exceeds GPU memory; quantized results only.}
  \end{tabular}
\end{table}
\vspace{-10pt}

\begin{table}[H]
  \centering\small
  \caption{Quantization impact on HellaSwag (10,042 examples, 0-shot). Focus on precision deltas within each model, not absolute scores.}
  \label{tab:api-quality}
  \begin{tabular}{l l l c}
    \toprule
    Model & Params (B) & Precision & HellaSwag Accuracy \\
    \midrule
    Qwen3-8B & 8.2 & NVFP4 & 0.7289 ± 0.0044 \\
    Qwen3-8B & 8.2 & BF16 (baseline) & 0.7491 ± 0.0043 \\
    Gemma3-12B$^\S$ & 12 & W4A16 & 0.3944 ± 0.0049 \\
    Gemma3-12B$^\S$ & 12 & NVFP4 & 0.3801 ± 0.0048  \\
    Gemma3-12B$^\S$ & 12 & BF16 (baseline) & 0.4009 ± 0.0049  \\
    Gemma3-27B$^{\ddagger,\S}$ & 27 & W4A16 & 0.4681 ± 0.0050 \\
    Gemma3-27B$^{\ddagger,\S}$ & 27 & NVFP4 &  0.4602 ± 0.0050 \\
    \bottomrule
    \multicolumn{4}{l}{\footnotesize $^\ddagger$BF16 baseline exceeds GPU memory; quantized results only.}\\
    \multicolumn{4}{l}{\footnotesize $^\S$0-shot eval; Google reports 84--86\% (10-shot, pre-trained).}
  \end{tabular}
\end{table}

\subsection{System-Level Performance (Summary)}

We benchmarked 79 valid configurations across three workload types (RAG, API, Agentic), four models, multiple quantization formats, and context lengths from 8k to 64k tokens. Detailed per-configuration results including throughput, latency percentiles, and energy consumption are provided in Appendix~\ref{app:detailed-results}. Table~\ref{tab:gpu-comparison} summarizes the key cross-GPU comparisons.

\begin{figure}[H]
  \centering
  \begin{tikzpicture}
    \begin{axis}[
      ybar,
      width=0.95\columnwidth,
      height=4.5cm,
      ylabel={Throughput (TPS)},
      symbolic x coords={RAG-8k, RAG-16k, API-c64},
      xtick=data,
      ymin=0,
      ymax=5500,
      legend style={at={(0.5,1.02)}, anchor=south, legend columns=4, font=\footnotesize},
      bar width=8pt,
      enlarge x limits=0.25,
      nodes near coords style={font=\tiny, rotate=90, anchor=west},
      every axis plot/.append style={fill opacity=0.85}
    ]
      \addplot[fill=gpu5060, draw=gpu5060!80!black] coordinates {(RAG-8k,115) (RAG-16k,52) (API-c64,1798)};
      \addplot[fill=gpu5070, draw=gpu5070!80!black] coordinates {(RAG-8k,211) (RAG-16k,97) (API-c64,2899)};
      \addplot[fill=gpu5090, draw=gpu5090!80!black] coordinates {(RAG-8k,411) (RAG-16k,232) (API-c64,4678)};
      \addplot[fill=gpu5090x2, draw=gpu5090x2!80!black] coordinates {(RAG-8k,530) (RAG-16k,303) (API-c64,4466)};
      \legend{5060 Ti 1x, 5070 Ti 1x, 5090 1x, 5090 2x}
    \end{axis}
  \end{tikzpicture}
  \caption{GPU performance hierarchy: throughput (TPS) for Qwen3-8B NVFP4 across workloads. The RTX 5090 delivers 3.5--4$\times$ higher throughput than the 5060 Ti.}
  \label{fig:gpu-hierarchy}
\end{figure}
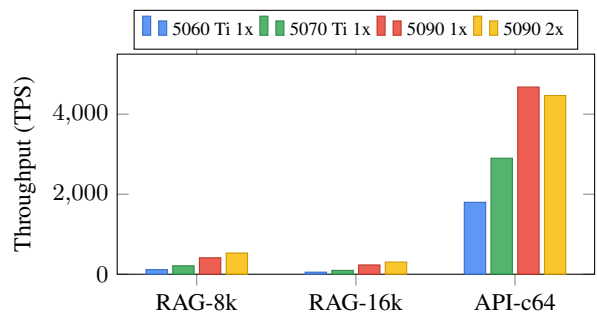

\begin{table}[H]
  \scriptsize
  \caption{Cross-GPU comparison summary. Best config.\ per workload in bold.}
  \label{tab:gpu-comparison}
  \begin{tabularx}{\columnwidth}{X X c c c c}
    \toprule
    Workload & Model/Prec. & 5060Ti & 5070Ti & 5090 & 5090×2 \\
    \midrule
    \multicolumn{6}{l}{\textit{Throughput (TPS)}} \\
    RAG-8k & Qwen3/NV4 & 115 & 211 & 411 & \textbf{530} \\
    RAG-16k & Qwen3/NV4 & 52 & 97 & 232 & \textbf{303} \\
    API-c64 & Qwen3/NV4 & 1798 & 2899 & \textbf{4678} & 4466 \\
    RAG-8k & Gem12/NV4 & -- & -- & 346 & \textbf{393} \\
    \midrule
    \multicolumn{6}{l}{\textit{TTFT (ms)}} \\
    RAG-8k & Qwen3/NV4 & 9658 & 5228 & \textbf{450} & 620 \\
    RAG-16k & Qwen3/NV4 & 12365 & 6503 & \textbf{1216} & 1478 \\
    API-c64 & Qwen3/NV4 & 351 & 237 & \textbf{131} & 173 \\
    \midrule
    \multicolumn{6}{l}{\textit{Energy (Wh/MTok)}} \\
    RAG-8k & Qwen3/NV4 & 298 & 275 & \textbf{239} & 330 \\
    API-c64 & Qwen3/NV4 & 20 & 19 & \textbf{18} & 32 \\
    \bottomrule
  \end{tabularx}
\end{table}

\FloatBarrier

We now address each research question posed in Section~\ref{sec:methods}.

\subsection{RQ1: Throughput and Latency Across GPU Tiers}

\textit{For the three workload classes, how do consumer Blackwell GPUs compare in terms of TPS, TTFT, and tail latencies across single- and dual-GPU configurations?}

\paragraph{Key Finding:} The RTX 5090 delivers 3.5--4.6$\times$ higher throughput than the RTX 5060 Ti across comparable workloads, with the performance gap widening dramatically for latency-sensitive applications.

\begin{table*}[!t]
  \centering\small
  \caption{Cross-GPU throughput comparison (Qwen3-8B NVFP4). RAG-8k at concurrency 8; RAG-16k at concurrency 8 (5090) or 4 (budget GPUs due to memory); API shows peak throughput.}
  \label{tab:rq1-throughput}
  \begin{tabular}{l l c c c c}
    \toprule
    Workload & Metric & 5060 Ti 1x & 5070 Ti 1x & 5090 1x & 5090 2x \\
    \midrule
    \multirow{2}{*}{RAG-8k} & TPS & 115 & 211 & 411 & \textbf{530} \\
     & TTFT (ms) & 9,658 & 5,228 & \textbf{450} & 620 \\
    \midrule
    \multirow{2}{*}{RAG-16k} & TPS & 52 & 97 & 232 & \textbf{303} \\
     & TTFT (ms) & 12,365 & 6,503 & \textbf{1,216} & 1,478 \\
    \midrule
    \multirow{2}{*}{API-peak} & TPS & 2,114 & 3,554 & 6,894 & \textbf{7,438} \\
     & TTFT (ms) & 620 & 361 & \textbf{177} & 600 \\
    \bottomrule
  \end{tabular}
\end{table*}

\paragraph{Latency Analysis.}
The RTX 5090 achieves sub-second TTFT (450ms) for 8k-context RAG workloads, compared to 5,228ms on the 5070 Ti and 9,658ms on the 5060 Ti---a 21$\times$ difference. For interactive applications requiring $<$1s response times, the RTX 5090 is the minimum viable option.

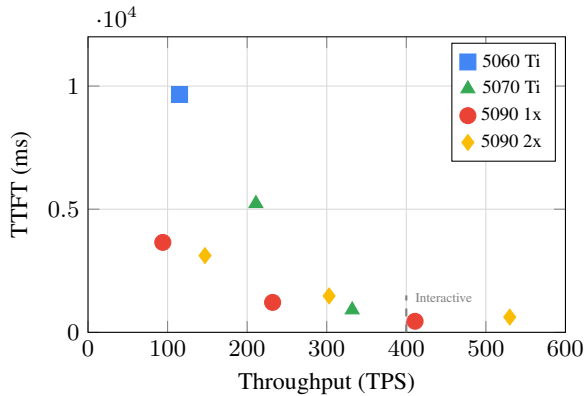
\begin{figure}[H]
  \centering
  \begin{tikzpicture}
    \begin{axis}[
      width=0.95\columnwidth,
      height=5.5cm,
      xlabel={Throughput (TPS)},
      ylabel={TTFT (ms)},
      xmin=0, xmax=600,
      ymin=0, ymax=12000,
      legend style={at={(0.98,0.98)}, anchor=north east, font=\footnotesize},
      grid=major,
      grid style={gray!30}
    ]
      \addplot[only marks, mark=square*, mark size=3pt, color=gpu5060] coordinates {(115,9658)};
      \addplot[only marks, mark=triangle*, mark size=3pt, color=gpu5070] coordinates {(211,5228) (332,912)};
      \addplot[only marks, mark=*, mark size=3pt, color=gpu5090] coordinates {(411,450) (232,1216) (94,3652)};
      \addplot[only marks, mark=diamond*, mark size=3pt, color=gpu5090x2] coordinates {(530,620) (303,1478) (147,3116)};
      \legend{5060 Ti, 5070 Ti, 5090 1x, 5090 2x}
      \draw[dashed, thick, gray] (axis cs:400,0) -- (axis cs:400,1500) node[pos=0.95, right, font=\tiny] {Interactive};
    \end{axis}
  \end{tikzpicture}
  \caption{Latency vs.\ throughput trade-off (Qwen3-8B NVFP4, RAG workloads). Lower-right is better. Only RTX 5090 configurations achieve sub-second TTFT with high throughput.}
  \label{fig:latency-throughput}
\end{figure}

\paragraph{Dual-GPU Scaling.}
Tensor parallelism provides 1.14--1.57$\times$ speedup depending on model size:

\begin{table*}[!t]
  \centering\small
  \caption{Tensor parallelism scaling efficiency (1x vs.\ 2x GPU, RAG-8k at concurrency 8).}
  \label{tab:rq1-tp-scaling}
  \begin{tabular}{l l c c c}
    \toprule
    Model & GPU Base & 1x TPS & 2x TPS & Speedup \\
    \midrule
    Qwen3-8B & RTX 5090 & 411 & 530 & 1.29$\times$ \\
    Qwen3-8B & RTX 5070 Ti & 211 & 332 & 1.57$\times$ \\
    Qwen3-8B & RTX 5060 Ti & 115 & 158 & 1.37$\times$ \\
    Gemma3-12B & RTX 5090 & 346 & 393 & 1.14$\times$ \\
    \bottomrule
  \end{tabular}
\end{table*}

The primary benefit of 2x configurations is extended context length support (32k$\rightarrow$64k+) rather than raw throughput scaling. Larger models (Gemma3-12B) show lower scaling efficiency (1.14$\times$) as they are already memory-bandwidth limited on a single GPU.

\paragraph{Tail Latencies.}
P95 latencies remain within 1.5--2$\times$ of median latencies for API workloads, indicating stable performance under load. RAG workloads show higher variance (P95/P50 = 1.3--1.8$\times$) due to context-dependent prefill times.

\begin{figure}[H]
  \centering
  \begin{tikzpicture}
    \begin{axis}[
      width=0.95\columnwidth,
      height=5cm,
      xlabel={Context Length (tokens)},
      ylabel={Throughput (TPS)},
      xmode=log,
      log basis x={2},
      xtick={8192, 16384, 32768, 65536},
      xticklabels={8k, 16k, 32k, 64k},
      ymin=0, ymax=600,
      legend style={at={(0.98,0.98)}, anchor=north east, font=\footnotesize},
      grid=major,
      grid style={gray!30}
    ]
      \addplot[thick, mark=square*, color=gpu5060] coordinates {(8192,115) (16384,52)};
      \addplot[thick, mark=triangle*, color=gpu5070] coordinates {(8192,211) (16384,97) (32768,34)};
      \addplot[thick, mark=*, color=gpu5090] coordinates {(8192,411) (16384,232) (32768,94)};
      \addplot[thick, mark=diamond*, color=gpu5090x2] coordinates {(8192,530) (16384,303) (32768,147)};
      \legend{5060 Ti 1x, 5070 Ti 1x, 5090 1x, 5090 2x}
    \end{axis}
  \end{tikzpicture}
  \caption{Context length impact on throughput (Qwen3-8B NVFP4). Doubling context approximately halves throughput.}
  \label{fig:context-scaling}
\end{figure}
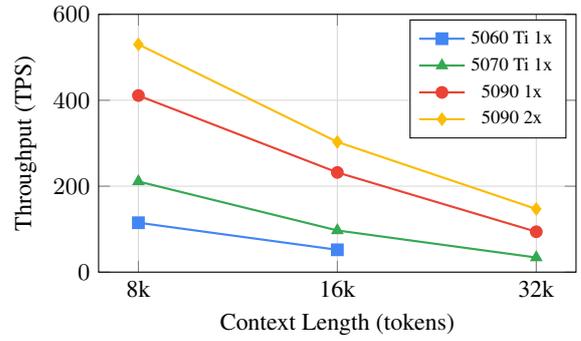

\subsection{RQ2: Quantization Trade-offs}

\textit{How do different low-precision formats---W4A16, NVFP4, and MXFP4---affect throughput, memory footprint, and energy per generated token?}

\paragraph{Key Finding:} NVFP4 provides the best performance-efficiency trade-off, delivering 1.6$\times$ throughput improvement over BF16 with 41\% energy reduction and minimal quality degradation.

\begin{table}[H]
  \centering\small
  \caption{Quantization format comparison (Qwen3-8B, 8k context, RTX 5090 1x, concurrency 8).}
  \label{tab:rq2-quant}
  \begin{tabular}{l c c c c}
    \toprule
    Format & TPS & TTFT (ms) & Wh/MTok & vs.\ BF16 \\
    \midrule
    BF16 (baseline) & 260 & 1,538 & 403 & 1.00$\times$ \\
    W4A16 & 314 & 1,030 & 325 & 1.21$\times$ \\
    NVFP4 & \textbf{411} & \textbf{450} & \textbf{239} & \textbf{1.58$\times$} \\
    \bottomrule
  \end{tabular}
\end{table}

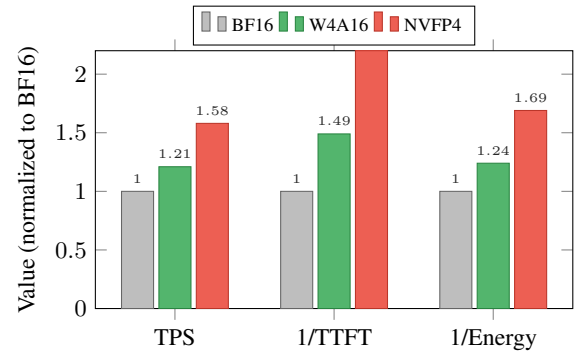
\begin{figure}[H]
  \centering
  \begin{tikzpicture}
    \begin{axis}[
      ybar,
      width=0.95\columnwidth,
      height=5cm,
      ylabel={Value (normalized to BF16)},
      symbolic x coords={TPS, 1/TTFT, 1/Energy},
      xtick=data,
      ymin=0, ymax=2.2,
      legend style={at={(0.5,1.02)}, anchor=south, legend columns=3, font=\footnotesize},
      bar width=12pt,
      enlarge x limits=0.25,
      nodes near coords,
      nodes near coords style={font=\tiny},
      every axis plot/.append style={fill opacity=0.85}
    ]
      \addplot[fill=gray!60, draw=gray!80!black] coordinates {(TPS,1.0) (1/TTFT,1.0) (1/Energy,1.0)};
      \addplot[fill=gpu5070, draw=gpu5070!80!black] coordinates {(TPS,1.21) (1/TTFT,1.49) (1/Energy,1.24)};
      \addplot[fill=gpu5090, draw=gpu5090!80!black] coordinates {(TPS,1.58) (1/TTFT,3.42) (1/Energy,1.69)};
      \legend{BF16, W4A16, NVFP4}
    \end{axis}
  \end{tikzpicture}
  \caption{Quantization format comparison (Qwen3-8B, RTX 5090 1x). Higher is better. NVFP4 outperforms W4A16 on all metrics.}
  \label{fig:quant-comparison}
\end{figure}

\paragraph{Throughput Gains.}
NVFP4 achieves:
\begin{itemize}
    \item 31\% higher throughput than W4A16 (411 vs.\ 314 TPS)
    \item 56\% lower TTFT than W4A16 (450ms vs.\ 1,030ms)
    \item 1.58$\times$ throughput vs.\ BF16 baseline
    \item 41\% lower energy consumption vs.\ BF16 (239 vs.\ 403 Wh/MTok)
\end{itemize}

\paragraph{Quality Preservation.}
From Tables~\ref{tab:rag-quality}--\ref{tab:api-quality}, quantization impacts are generally acceptable:
\begin{itemize}
    \item Qwen3-8B: NVFP4 vs.\ BF16 shows $-$2.2\% on MMLU, $-$1.9\% on GSM8K, $-$2.0\% on HellaSwag
    \item Gemma3-12B: NVFP4 vs.\ BF16 shows $-$4.1\% on MMLU, $-$3.6\% on GSM8K (slightly above our 1--2\% target)
    \item W4A16 generally preserves quality better than NVFP4 but with lower throughput gains
\end{itemize}

\paragraph{Memory Efficiency.}
NVFP4 reduces memory footprint by approximately 50\% compared to BF16, enabling:
\begin{itemize}
    \item Longer context lengths on the same GPU (32k$\rightarrow$64k on RTX 5090)
    \item Larger models on smaller GPUs (Gemma3-12B on RTX 5060 Ti 2x)
    \item Higher concurrent request capacity
\end{itemize}

\paragraph{MXFP4 for MoE Models.}
GPT-OSS-20B with MXFP4 achieves 319--424 TPS on RTX 5090 1x at 8k context, competitive with dense models despite its larger parameter count. Context scaling follows expected patterns: throughput decreases to 203 TPS at 16k and 100 TPS at 32k on single GPU. Dual-GPU configurations extend capabilities significantly, achieving 501 TPS at 16k context and enabling 64k context (40 TPS) that is infeasible on single GPU. Notably, for short-context API workloads (256 tokens), GPT-OSS-20B achieves 488 TPS even on RTX 5060 Ti 1x, demonstrating that the sparse MoE architecture combined with 4-bit quantization enables efficient deployment on 16GB GPUs when context requirements are modest.

\subsection{RQ3: Agentic Multi-LoRA Overheads}

\textit{What is the overhead of frequent LoRA adapter switching, and how well does vLLM's adapter management scale?}

\paragraph{Key Finding:} vLLM's LoRA adapter management introduces minimal overhead. Throughput scales efficiently with concurrency, and adapter switching does not cause significant performance degradation.

\begin{table*}[!t]
  \centering\small
  \caption{Agentic workload performance (Qwen3-8B NVFP4, 2k context, 3 LoRA adapters).}
  \label{tab:rq3-agentic}
  \begin{tabular}{l l c c c c}
    \toprule
    GPU Config & Concurrency & TPS & TPS/User & TTFT (ms) & Wh/MTok \\
    \midrule
    RTX 5090 1x & 16 & 889 & 57.5 & 116 & 82.6 \\
    RTX 5090 1x & 32 & 1,304 & 42.7 & 198 & 64.2 \\
    RTX 5090 1x & 64 & \textbf{1,683} & 28.1 & 412 & \textbf{53.7} \\
    RTX 5090 2x & 32 & 1,176 & 36.8 & 672 & 100 \\
    RTX 5090 2x & 64 & 1,492 & 23.3 & 1,579 & 82 \\
    \midrule
    RTX 5070 Ti 1x & 8 & 442 & 57.5 & 191 & 140.6 \\
    RTX 5070 Ti 1x & 16 & 661 & 42.8 & 227 & 90 \\
    RTX 5070 Ti 1x & 32 & 677 & 35.3 & 7,004 & 87 \\
    RTX 5070 Ti 1x & 64 & 1,044 & 16.3 & 11,634 & 69 \\
    RTX 5070 Ti 2x & 32 & 833 & 28.2 & 604 & 224 \\
    \midrule
    RTX 5060 Ti 1x & 8 & 264 & 34.2 & 319 & 145.7 \\
    RTX 5060 Ti 1x & 16 & 366 & 23.8 & 423 & 104.6 \\
    RTX 5060 Ti 1x & 32 & 409 & 19.5 & 9,868 & 96.7 \\
    RTX 5060 Ti 1x & 64 & 411 & 19.5 & 37,077 & 95.5 \\
    RTX 5060 Ti 2x & 32 & 506 & 17.1 & 998 & 219 \\
    RTX 5060 Ti 2x & 64 & 546 & 9.4 & 2,502 & 200 \\
    \bottomrule
  \end{tabular}
\end{table*}

\begin{figure}[H]
  \centering
  \begin{tikzpicture}
    \begin{axis}[
      width=0.95\columnwidth,
      height=5cm,
      xlabel={Concurrency},
      ylabel={Throughput (TPS)},
      xmin=0, xmax=70,
      ymin=0, ymax=1800,
      legend style={at={(0.02,0.98)}, anchor=north west, font=\footnotesize},
      grid=major,
      grid style={gray!30}
    ]
      \addplot[thick, mark=*, color=gpu5090] coordinates {(16,889) (32,1304) (64,1683)};
      \addplot[thick, mark=triangle*, color=gpu5070] coordinates {(8,442) (16,661) (32,677) (64,1044)};
      \addplot[thick, mark=square*, color=gpu5060] coordinates {(8,264) (16,366) (32,409) (64,411)};
      \legend{RTX 5090 1x, RTX 5070 Ti 1x, RTX 5060 Ti 1x}
      \draw[dashed, gray] (axis cs:32,409) -- (axis cs:64,411);
      \node[font=\tiny, anchor=west] at (axis cs:50,450) {Saturated};
    \end{axis}
  \end{tikzpicture}
  \caption{Agentic workload scaling (Qwen3-8B NVFP4, 3 LoRA adapters). RTX 5090 and 5070 Ti scale with concurrency; RTX 5060 Ti saturates at c32.}
  \label{fig:agentic-scaling}
\end{figure}
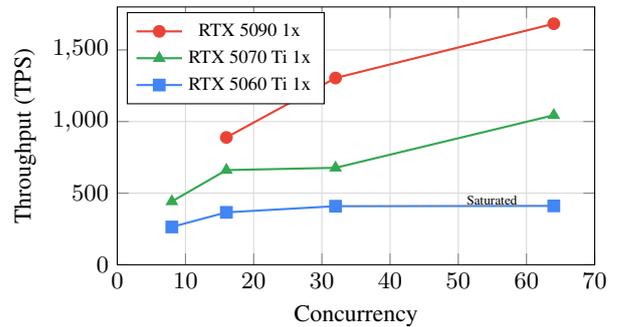

\paragraph{Scaling Behavior.}
\begin{itemize}
    \item RTX 5090 1x: Throughput scales from 889 TPS (c16) to 1,683 TPS (c64), a 1.89$\times$ improvement
    \item RTX 5070 Ti 1x: Scales from 442 TPS (c8) to 1,044 TPS (c64), with continued scaling at high concurrency; dual-GPU (2x) achieves 833 TPS at c32
    \item RTX 5060 Ti 1x: Throughput plateaus at 409--411 TPS beyond c32, indicating GPU saturation; dual-GPU (2x) extends to 546 TPS at c64
    \item TTFT increases with concurrency as expected, but remains $<$500ms on RTX 5090 at c64
    \item \textbf{RTX 5090 2x underperforms 1x:} Counter-intuitively, dual-GPU achieves lower throughput (1,492 vs.\ 1,683 TPS at c64) with 3.8$\times$ higher latency (1,579ms vs.\ 412ms). This results from tensor-parallelism communication overhead compounding with LoRA adapter synchronization across GPUs; for short-context workloads (2k tokens), inter-GPU coordination costs exceed parallelism benefits---a pattern not observed in longer-context RAG workloads where compute dominates communication
\end{itemize}

\paragraph{Adapter Switching Overhead.}
Comparing agentic (multi-LoRA) to single-model API workloads at similar concurrency:
\begin{itemize}
    \item RTX 5090 1x at c64: Agentic 1,683 TPS vs.\ API 4,678 TPS (2.8$\times$ ratio)
    \item The difference is primarily due to longer context (2k vs.\ 256 tokens), not adapter overhead
    \item Per-user TPS remains stable across adapters, indicating efficient adapter management
\end{itemize}

\paragraph{Recommendations for Multi-Tenant Deployment.}
\begin{itemize}
    \item RTX 5090 \textbf{single-GPU} is optimal for responsive multi-LoRA serving (TTFT $<$500ms at c64); 2x configurations hurt performance due to synchronization overhead
    \item Budget GPUs (5060 Ti, 5070 Ti) benefit from 2x for agentic workloads: dual-GPU reduces TTFT by 11--15$\times$ while increasing throughput
    \item vLLM's adapter caching effectively eliminates switching overhead for our tested pool of 3 adapters
\end{itemize}

\subsection{RQ4: Energy and Cost Efficiency}

\textit{What is the energy consumption and electricity cost per million tokens across configurations? Which combinations offer the best energy efficiency?}

\paragraph{Key Finding:} Energy efficiency varies by 100$\times$ across workloads and configurations. API workloads achieve \$0.001--0.005/MTok, while RAG-32k costs \$0.14--0.22/MTok.

\begin{figure}[H]
  \centering
  \begin{tikzpicture}
    \begin{axis}[
      xbar,
      width=0.90\columnwidth,
      height=5cm,
      xlabel={Energy (Wh/MTok)},
      symbolic y coords={API-c128, Agentic-c16, RAG-8k, RAG-32k},
      ytick=data,
      xmin=0, xmax=1200,
      legend style={at={(0.98,0.02)}, anchor=south east, font=\footnotesize},
      bar width=8pt,
      enlarge y limits=0.15,
      nodes near coords,
      nodes near coords style={font=\tiny, anchor=west},
      every axis plot/.append style={fill opacity=0.85}
    ]
      \addplot[fill=gpu5090, draw=gpu5090!80!black] coordinates {(13,API-c128) (83,Agentic-c16) (239,RAG-8k) (1146,RAG-32k)};
      \addplot[fill=gpu5060, draw=gpu5060!80!black] coordinates {(17,API-c128) (105,Agentic-c16) (298,RAG-8k)};
      \legend{RTX 5090 1x, RTX 5060 Ti 1x}
    \end{axis}
  \end{tikzpicture}
  \caption{Energy consumption by workload (Qwen3-8B NVFP4). API workloads are 18--88$\times$ more efficient than long-context RAG.}
  \label{fig:energy-efficiency}
\end{figure}
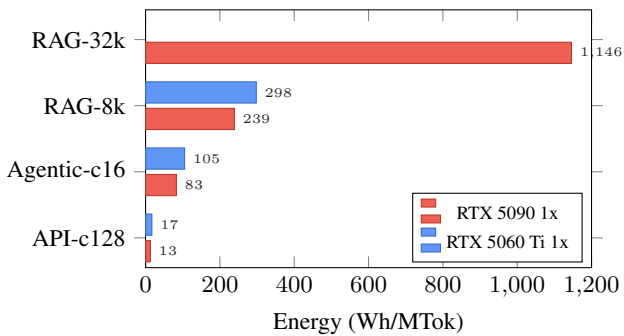

\begin{table*}[!t]
  \centering\small
  \caption{Energy efficiency by workload type (\$0.12/kWh electricity rate).}
  \label{tab:rq4-energy}
  \begin{tabular}{l l l c c c}
    \toprule
    Workload & Model & GPU Config & Wh/MTok & \$/MTok & Power (W) \\
    \midrule
    \multicolumn{6}{l}{\textit{API Workloads (Best Efficiency)}} \\
    API-c256 & Qwen3-8B & RTX 5090 2x & 20.4 & 0.0024 & 546 \\
    API-c128 & Qwen3-8B & RTX 5090 1x & \textbf{12.6} & \textbf{0.0015} & 309 \\
    API-c128 & Qwen3-8B & RTX 5070 Ti 1x & 15.8 & 0.0019 & 202 \\
    API-c128 & Qwen3-8B & RTX 5060 Ti 1x & 16.9 & 0.0020 & 128 \\
    \midrule
    \multicolumn{6}{l}{\textit{RAG-8k Workloads}} \\
    RAG-8k & Qwen3-8B & RTX 5090 1x & \textbf{239} & \textbf{0.029} & 353 \\
    RAG-8k & Qwen3-8B & RTX 5070 Ti 1x & 275 & 0.033 & 208 \\
    RAG-8k & Qwen3-8B & RTX 5060 Ti 1x & 298 & 0.036 & 124 \\
    RAG-8k & Qwen3-8B & RTX 5090 2x & 330 & 0.040 & 629 \\
    \midrule
    \multicolumn{6}{l}{\textit{RAG-32k Workloads (Highest Cost)}} \\
    RAG-32k & Qwen3-8B & RTX 5090 1x & \textbf{1,146} & \textbf{0.138} & 390 \\
    RAG-32k & Qwen3-8B & RTX 5090 2x & 1,531 & 0.184 & 811 \\
    RAG-32k & Qwen3-8B & RTX 5060 Ti 2x & 1,828 & 0.219 & 234 \\
    \bottomrule
  \end{tabular}
\end{table*}

\paragraph{Context Length Impact.}
Doubling context length approximately halves throughput and doubles energy cost per token:
\begin{itemize}
    \item 8k$\rightarrow$16k: 1.8--2.3$\times$ energy increase
    \item 16k$\rightarrow$32k: 2.0--2.5$\times$ energy increase
    \item 32k context costs 5.5$\times$ the 8k baseline
\end{itemize}

\paragraph{Comparison to Cloud APIs.}
Self-hosted inference provides substantially lower per-token costs compared to commercial API providers. As of January 2026, we compare against four representative cloud models: budget-tier options GPT-5 nano (\$0.05/\$0.40 input/output), Gemini 2.0 Flash-Lite (\$0.075/\$0.30), and GPT-5 mini (\$0.25/\$2.00), plus frontier-class Claude Opus 4.5 (\$5/\$25)~\cite{openai2025pricing,google2025pricing,anthropic2025pricing}. Using blended rates (assuming equal input/output volumes), these translate to approximately \$0.19/MTok (Gemini Flash-Lite), \$0.23/MTok (GPT-5 nano), \$1.13/MTok (GPT-5 mini), and \$15/MTok (Claude Opus 4.5). The budget-tier models offer comparable capability to the 8--27B open models we benchmark; Opus 4.5 represents frontier performance.\footnote{GPT-5 nano/mini and Gemini Flash-Lite target similar use cases as Qwen3-8B and Gemma3-12B (fast inference for chatbots, classification, summarization); Gemma3-27B and GPT-OSS-20B approach GPT-5 mini capability for more complex tasks.} Our measured self-hosted costs of \$0.001--0.005/MTok for API workloads represent a 40--230$\times$ reduction compared to budget-tier cloud APIs and 3,000--15,000$\times$ compared to frontier models.

\paragraph{Break-Even Analysis.}
We compute break-even time $T_{\text{BE}}$ as the point where cumulative cloud API costs equal hardware investment plus electricity:
\begin{equation}
  T_{\text{BE}} = \frac{C_{\text{hw}}}{V_{\text{daily}} \times (P_{\text{cloud}} - P_{\text{self}})}
\end{equation}
where $C_{\text{hw}}$ is hardware cost, $V_{\text{daily}}$ is daily token volume, $P_{\text{cloud}}$ and $P_{\text{self}}$ are per-token costs.

For an RTX 5090 (\$2,000) at 30M tokens/day vs.\ GPT-5 nano (\$0.23/MTok):
\begin{equation}
  T_{\text{BE}} = \frac{\$2{,}000}{\$6.84/\text{day}} \approx \textbf{292 days}
\end{equation}

Against Claude Opus 4.5 (\$15/MTok) at 30M tokens/day:
\begin{equation}
  T_{\text{BE}} = \frac{\$2{,}000}{\$450/\text{day}} \approx \textbf{4 days}
\end{equation}

At lower volumes (1M tokens/day), break-even periods extend proportionally (e.g., 8,772 days for RTX 5090 vs.\ GPT-5 nano). Table~\ref{tab:breakeven} summarizes break-even periods at 30M tokens/day, representative of a small team or department with moderate AI usage.

\begin{table*}[!t]
  \centering\small
  \caption{Break-even analysis: days until self-hosted costs equal cloud API investment (January 2026 pricing).}
  \label{tab:breakeven}
  \begin{tabular}{l c c c c}
    \toprule
    GPU Config & Cost & \multicolumn{3}{c}{Break-even (days) at 30M tok/day} \\
    \cmidrule(lr){3-5}
    & & vs.\ GPT-5 nano & vs.\ GPT-5 mini & vs.\ Claude Opus 4.5 \\
    \midrule
    RTX 5060 Ti 1x & \$500 & 73 & 15 & 1 \\
    RTX 5070 Ti 1x & \$900 & 132 & 27 & 2 \\
    RTX 5090 1x & \$2,000 & 292 & 59 & 4 \\
    RTX 5090 2x & \$4,000 & 585 & 118 & 9 \\
    \bottomrule
  \end{tabular}
\end{table*}

These break-even estimates use API workload costs (\$0.001--0.005/MTok), representing the best-case scenario. RAG workloads incur higher per-token costs (\$0.03/MTok for 8k context, \$0.14/MTok for 32k), extending break-even periods by 5--30$\times$. Even so, self-hosted RAG remains cost-competitive with budget-tier cloud APIs while providing complete data sovereignty.

\begin{figure}[H]
  \centering
  \begin{tikzpicture}
    \begin{axis}[
      ybar,
      width=0.95\columnwidth,
      height=6cm,
      ylabel={Cost (\$/MTok)},
      symbolic x coords={Self-hosted API, Self-hosted RAG-8k, Gemini Flash-Lite, GPT-5 nano, GPT-5 mini, Claude Opus 4.5},
      xtick=data,
      x tick label style={rotate=35, anchor=east, font=\footnotesize},
      ymin=0, ymax=18,
      bar width=12pt,
      nodes near coords,
      nodes near coords style={font=\tiny, anchor=south},
      every axis plot/.append style={fill opacity=0.85}
    ]
      \addplot[fill=gpu5090, draw=gpu5090!80!black] coordinates {(Self-hosted API,0.002) (Self-hosted RAG-8k,0.03) (Gemini Flash-Lite,0.19) (GPT-5 nano,0.23) (GPT-5 mini,1.13) (Claude Opus 4.5,15)};
    \end{axis}
  \end{tikzpicture}
  \caption{Cost comparison: self-hosted inference (RTX 5090, NVFP4) vs.\ commercial APIs (January 2026 pricing). Self-hosted costs represent electricity only; hardware amortization adds \$0.01--0.05/MTok over 2 years.}
  \label{fig:cost-comparison}
\end{figure}
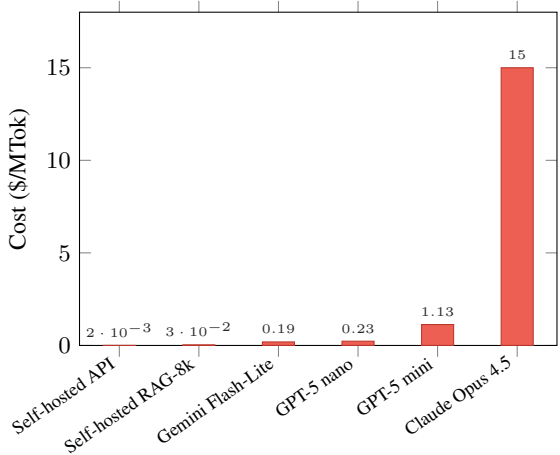

\paragraph{Optimal Configurations by Workload.}
Based on our benchmarks, we identify workload-specific configurations that optimize the cost-performance trade-off:

\textbf{High-Throughput API/Chatbot Workloads.} For applications requiring maximum throughput with short contexts (customer service bots, coding assistants), RTX 5090 2x with Qwen3-8B NVFP4 at c128--c256 concurrency achieves 4,400+ TPS at \$0.002/MTok. This configuration sustains 380M tokens/day at under \$800/year electricity cost, making it suitable for high-volume production deployments.

\textbf{RAG Workloads (8k--16k context).} Retrieval-augmented generation with moderate context lengths benefits from RTX 5090 1x configurations for cost efficiency. At 8k context, Qwen3-8B NVFP4 delivers 411 TPS at \$0.029/MTok (1x) or 530 TPS at \$0.040/MTok (2x). For organizations processing 10M RAG queries monthly on a single RTX 5090, this translates to approximately \$290/month in electricity versus \$1,900--15,000/month for equivalent cloud API usage (Gemini Flash-Lite to GPT-5 mini).

\textbf{Long-Context RAG (32k+).} Extended context workloads require dual-GPU configurations to maintain acceptable latency. RTX 5090 2x achieves 303 TPS at 32k context with TTFT under 2 seconds, compared to 10+ second TTFT on single-GPU configurations. The 5.5$\times$ cost increase from 8k to 32k context makes semantic chunking strategies economically compelling for borderline use cases.

\textbf{Budget-Constrained Deployments.} The RTX 5060 Ti and 5070 Ti offer compelling value propositions for specific use cases, particularly short-context API workloads where latency requirements are less stringent than RAG applications.

\begin{table*}[!t]
  \centering\small
  \caption{Cost-efficiency comparison across GPU tiers (Qwen3-8B NVFP4, API-c128 workload).}
  \label{tab:cost-efficiency}
  \begin{tabular}{l c c c c c}
    \toprule
    GPU & HW Cost & TPS & TTFT (ms) & TPS/\$1k & Wh/MTok \\
    \midrule
    RTX 5060 Ti 1x & \$500 & 2,114 & 620 & \textbf{4,228} & 16.9 \\
    RTX 5070 Ti 1x & \$900 & 3,554 & 361 & 3,949 & 15.8 \\
    RTX 5090 1x & \$2,000 & 6,809 & 206 & 3,404 & 12.6 \\
    \bottomrule
  \end{tabular}
\end{table*}

The RTX 5060 Ti achieves the highest throughput-per-dollar ratio (4,228 TPS/\$1k) for API workloads, making it the most cost-efficient option when absolute performance is not the primary constraint. Critically, the 620ms TTFT for API workloads is substantially lower than RAG workloads (9,658ms at 8k context) and remains acceptable for many web applications where users tolerate 1--2 second response times.

\paragraph{Recommended Use Cases for Budget GPUs.}
Based on our benchmarks, we identify specific scenarios where RTX 5060 Ti or 5070 Ti represent the optimal choice rather than merely an acceptable compromise:

\begin{itemize}
    \item \textbf{Internal chatbots (5--15 concurrent users)}: RTX 5060 Ti at c16 delivers 366 TPS with 23 TPS/user and 423ms TTFT---sufficient for internal productivity tools, customer service training, or document Q\&A systems where sub-second response is not critical.
    \item \textbf{Development and staging environments}: RTX 5060 Ti provides full API compatibility with production 5090 deployments at 25\% of the hardware cost, enabling realistic load testing and model evaluation.
    \item \textbf{Moderate-traffic APIs ($<$50 requests/second)}: RTX 5070 Ti handles 3,554 TPS with 361ms TTFT, suitable for SaaS products serving small-to-medium customer bases where response times under 500ms are acceptable.
    \item \textbf{Multi-LoRA for small teams}: RTX 5070 Ti at c8 achieves 56 TPS/user with 185ms TTFT across 3 LoRA adapters, enabling personalized assistants for teams of 5--10 users with responsive performance.
    \item \textbf{Batch processing and offline inference}: When latency is irrelevant (document summarization, data extraction, content generation pipelines), RTX 5060 Ti maximizes tokens-per-dollar at \$0.002/MTok electricity cost.
    \item \textbf{MoE model deployment}: GPT-OSS-20B with MXFP4 achieves 488 TPS on RTX 5060 Ti for short-context API workloads, demonstrating that 20B-parameter sparse models are deployable on 16GB consumer GPUs.
\end{itemize}

\paragraph{Dual-GPU Budget Configurations (2x).}
Tensor parallelism on budget GPUs provides substantial benefits beyond raw throughput scaling:

\begin{table}[H]
  \small
  \caption{Benefits of dual-GPU configurations on budget hardware (Qwen3-8B NVFP4).}
  \label{tab:budget-2x}
  \begin{tabularx}{\columnwidth}{X X c c c c}
    \toprule
    GPU Config & Workload & \multicolumn{2}{c}{TPS} & \multicolumn{2}{c}{TTFT (ms)} \\
    \cmidrule(lr){3-4} \cmidrule(lr){5-6}
    & & 1x & 2x & 1x & 2x \\
    \midrule
    RTX 5070 Ti & Agentic-c32 & 677 & 833 & 7,004 & \textbf{604} \\
    RTX 5070 Ti & RAG-8k-c8 & 211 & 332 & 5,228 & \textbf{912} \\
    RTX 5060 Ti & Agentic-c32 & 409 & 506 & 9,868 & \textbf{998} \\
    RTX 5060 Ti & Agentic-c64 & 411 & 546 & 37,077 & \textbf{2,502} \\
    \bottomrule
  \end{tabularx}
\end{table}

The primary benefit is \textit{latency reduction}, not throughput scaling. RTX 5070 Ti 2x reduces Agentic-c32 TTFT by 11$\times$ (7s$\rightarrow$604ms), transforming an unusable configuration into an interactive one. Additionally, 2x configurations enable:
\begin{itemize}
    \item \textbf{Larger models on budget hardware}: Gemma3-27B achieves 58 TPS (W4A16) on RTX 5070 Ti 2x and 39 TPS on RTX 5060 Ti 2x---previously impossible on single 16GB GPUs.
    \item \textbf{Extended context}: RAG-32k becomes viable on RTX 5070 Ti 2x (74 TPS) where single-GPU configurations fail or exhibit unacceptable latency.
    \item \textbf{Higher concurrency}: RTX 5060 Ti 2x handles 64 concurrent agentic users with 2.5s TTFT versus 37s on single GPU.
\end{itemize}

At \$1,000 for dual RTX 5060 Ti or \$1,800 for dual RTX 5070 Ti, these configurations offer compelling alternatives to a single RTX 5090 (\$2,000) for workloads prioritizing model capability or context length over absolute latency.

\paragraph{When RTX 5090 is Essential.}
The higher-tier GPU remains necessary for: (1) interactive applications requiring $<$200ms TTFT, (2) single-GPU simplicity with sub-second RAG latency, and (3) highest throughput production serving ($>$100 concurrent users).

\subsection{Practical Recommendations for End Users}

Based on our findings, we provide deployment guidance for SMEs:

\subsubsection{GPU Selection Guide}

\begin{table}[H]
  \small
  \caption{GPU selection by use case and budget (TTFT varies by workload).}
  \label{tab:gpu-guide}
  \begin{tabularx}{\columnwidth}{l X c c l}
    \toprule
    GPU & Best For & \multicolumn{2}{c}{TTFT (ms)} & Cost/MTok \\
    \cmidrule(lr){3-4}
    & & API & RAG-8k & \\
    \midrule
    RTX 5090 1x & Production, interactive & 206 & 450 & \$0.02--0.03 \\
    RTX 5090 2x & High-volume, long context & 284 & 620 & \$0.02--0.04 \\
    RTX 5070 Ti 1x & Moderate traffic, small teams & 361 & 5,228 & \$0.03--0.04 \\
    RTX 5060 Ti 1x & Internal tools, batch, dev/test & 620 & 9,658 & \$0.03--0.04 \\
    \bottomrule
  \end{tabularx}
\end{table}

\paragraph{Key Insight.} API workloads (short context, 256 tokens) achieve sub-second TTFT on \textit{all} GPU tiers, making budget GPUs viable for chatbots and coding assistants. RAG workloads (8k+ context) show 10--20$\times$ higher latency on budget GPUs, restricting them to batch or latency-tolerant applications.

\subsubsection{Model and Quantization Selection}

\begin{itemize}
    \item \textbf{Default choice}: Qwen3-8B NVFP4---best throughput-to-quality ratio, fits on all tested GPUs
    \item \textbf{Higher capability}: Gemma3-12B NVFP4---larger model with broader factual knowledge, 15\% lower throughput. \textit{Important}: requires RTX 5090 or dual-GPU configuration for RAG workloads; single 16GB GPUs support short-context API only
    \item \textbf{Maximum capability}: Gemma3-27B NVFP4---largest dense model for complex tasks, requires RTX 5090 (1x or 2x), 3$\times$ lower throughput
    \item \textbf{MoE efficiency}: GPT-OSS-20B MXFP4---sparse architecture enables 488 TPS on RTX 5060 Ti for short-context API workloads, competitive with smaller dense models; longer contexts (RAG) require RTX 5090
\end{itemize}

\subsubsection{Context Length Strategy}

Context length is the primary cost driver. We recommend:
\begin{itemize}
    \item Prefer 8k context where possible (lowest cost)
    \item Use semantic chunking to stay within 16k for most RAG applications
    \item Reserve 32k+ for tasks that genuinely require long context
    \item Consider dual-GPU configurations for consistent 32k+ workloads
\end{itemize}

\section{Limitations}
\label{sec:limitations}

Several limitations constrain generalizability. Our evaluation focuses on NVIDIA consumer GPUs; while NVFP4 is Blackwell-specific, similar FP4/MXFP4 formats are emerging across vendors (AMD, Intel), and our methodology transfers directly to comparable hardware. NVFP4 requires calibration via official checkpoints or self-quantization (as we did for Gemma3 using llm-compressor). Our quality evaluation uses three standard benchmarks (MMLU, GSM8K, HellaSwag), which may not capture domain-specific degradation patterns. Our cost analysis uses manufacturer MSRP; actual street prices may vary due to supply constraints, and practitioners should adjust break-even calculations accordingly.

Experimentally, long-context benchmarks beyond 64k remain incomplete due to memory constraints. Several configurations failed with OOM errors: Gemma3-12B exceeds VRAM on single 16GB GPUs for RAG workloads, requiring RTX 5090 or dual-GPU setups. The agentic evaluation used only three LoRA adapters; production deployments may involve dozens with different access patterns.

Methodologically, we measure steady-state performance under synthetic workloads; production exhibits variable request patterns and cold-start effects. Energy measurements via DCGM capture GPU power only, excluding CPU/memory/cooling overhead (20--40\% additional); total system energy would increase absolute values but not relative ordering across configurations. All benchmarks use vLLM; alternative engines (TensorRT-LLM, SGLang, llama.cpp) may yield different characteristics.

\section{Conclusion}
\label{sec:conclusion}

For SMEs evaluating local LLM deployment on consumer GPUs, we offer the following key takeaways:
\begin{enumerate}
    \item \textbf{GPU selection depends on workload type.} RTX 5090 is required for interactive RAG applications ($<$1s TTFT), but RTX 5060 Ti is viable for API/chatbot workloads (620ms TTFT) and batch processing at 4$\times$ better cost efficiency.
    \item \textbf{NVFP4 should be the default quantization choice on Blackwell GPUs.} It delivers 1.6$\times$ throughput over BF16 and 31\% over W4A16 (the current local inference standard) with 2--4\% quality degradation on standard benchmarks.
    \item \textbf{Context length is the primary cost driver.} RAG-32k costs 5.5$\times$ more than RAG-8k; prefer semantic chunking to stay within 16k where possible.
    \item \textbf{Dual-GPU benefits are workload-dependent.} For RAG, 2x reduces TTFT by 11$\times$ on budget GPUs, transforming unusable configurations into interactive ones. For API, single-GPU suffices with better latency (206ms vs 600ms on RTX 5090). For agentic workloads, budget GPUs benefit from 2x, but RTX 5090 performs better with single-GPU due to synchronization overhead.
    \item \textbf{Self-hosted inference offers 40--200$\times$ cost reduction} versus budget-tier cloud APIs for typical workloads, with break-even periods of 15--118 days at moderate volume (30M tokens/day vs.\ GPT-5 mini).
    \item \textbf{Start with Qwen3-8B NVFP4.} It fits on all tested GPUs, offers the best throughput-to-quality ratio, and serves as a reliable baseline before scaling to larger models.
\end{enumerate}

\begin{table}[H]
  \small
  \caption{GPU configuration decision matrix. \textbf{Bold} = recommended.}
  \label{tab:decision-matrix}
  \begin{tabularx}{\columnwidth}{l XXX}
    \toprule
    & RTX 5090 & RTX 5070 Ti & RTX 5060 Ti \\
    \midrule
    \textbf{RAG-8k} & & & \\
    \quad Config & \textbf{1x} & 2x required & \textbf{2x (batch)} \\
    \quad TPS & 411 & 332 (2x) & 158 (2x) \\
    \quad TTFT & \textbf{450ms} & 912ms & 2.6s \\
    \midrule
    \textbf{API} & & & \\
    \quad Config & \textbf{1x preferred} & \textbf{1x sufficient} & \textbf{1x best value} \\
    \quad TPS & 4.7--6.8k & 2.9--3.6k & 1.8--2.1k \\
    \quad TTFT & \textbf{131--206ms} & 237--361ms & \textbf{$<$1s} \\
    \midrule
    \textbf{Agentic} & & & \\
    \quad Config & \textbf{1x preferred} & 1x / 2x & \textbf{2x (budget)} \\
    \quad TPS & 889--1.7k & 447--661 & 264--366 \\
    \quad TTFT & \textbf{115--412ms} & 185--227ms & 319--423ms \\
    \bottomrule
  \end{tabularx}
\end{table}

\clearpage
\onecolumn
\appendix
\section{Detailed Benchmark Results}
\label{app:detailed-results}

\begin{table}[H]
\centering\footnotesize
\caption{RAG workload results across all GPU configurations.}
\label{tab:rag-throughput}
\begin{tabularx}{\textwidth}{X l l c c c c c c}
  \toprule
  Model & GPU Config & Precision & Context & Conc. & TPS & TTFT & P95 & P99 \\
    \midrule
    \multicolumn{9}{l}{\textit{RTX 5090}} \\
    Qwen3-8B & 1x & NVFP4 & 8k & 8 & 410.7 & 450 & 1091 & 1371 \\
    Qwen3-8B & 1x & NVFP4 & 16k & 8 & 231.5 & 1216 & 2681 & 3236 \\
    Qwen3-8B & 1x & NVFP4 & 32k & 4 & 94.2 & 3652 & 4757 & 8101 \\
    Gemma3-12B & 1x & W4A16 & 8k & 4 & 206.0 & 1102 & 1846 & 2068 \\
    Gemma3-12B & 1x & NVFP4 & 8k & 8 & 345.9 & 944 & 1652 & 1708 \\
    Gemma3-27B & 1x & W4A16 & 8k & 4 & 111.6 & 6817 & 9087 & 9101 \\
    Gemma3-27B & 1x & NVFP4 & 8k & 4 & 98.5 & 5722 & 14223 & 14291 \\
    Gemma3-27B & 1x & NVFP4 & 32k & 4 & 31.4 & 50227 & 52321 & 53081 \\
    GPT-OSS-20B & 1x & MXFP4 & 8k & 4 & 319.5 & 519 & 925 & 960 \\
    GPT-OSS-20B & 1x & MXFP4 & 32k & 4 & 100.5 & 3211 & 5626 & 5698 \\
    Qwen3-8B & 2x & NVFP4 & 8k & 8 & 530.1 & 620 & 1800 & 2794 \\
    Qwen3-8B & 2x & NVFP4 & 16k & 8 & 303.3 & 1478 & 4235 & 5020 \\
    Qwen3-8B & 2x & NVFP4 & 32k & 8 & 147.0 & 3116 & 6532 & 8877 \\
    Gemma3-12B & 2x & NVFP4 & 8k & 8 & 393.2 & 1414 & 2638 & 3434 \\
    Gemma3-27B & 2x & NVFP4 & 32k & 4 & 23.5 & 20051 & 36201 & 37020 \\
    GPT-OSS-20B & 2x & MXFP4 & 16k & 8 & 500.8 & 806 & 4163 & 10785 \\
    GPT-OSS-20B & 2x & MXFP4 & 64k & 4 & 40.4 & 11846 & 19728 & 27230 \\
    \midrule
    \multicolumn{9}{l}{\textit{RTX 5070 Ti}} \\
    Qwen3-8B & 1x & NVFP4 & 8k & 8 & 210.8 & 5228 & 9963 & 12194 \\
    Qwen3-8B & 1x & NVFP4 & 16k & 4 & 96.7 & 6503 & 11608 & 13249 \\
    Qwen3-8B & 1x & NVFP4 & 32k & 4 & 33.9 & 46095 & 48477 & 48560 \\
    Gemma3-12B & 1x & W4A16 & 8k & -- & X & X & X & X \\
    Gemma3-12B & 1x & NVFP4 & 8k & -- & X & X & X & X \\
    Qwen3-8B & 2x & NVFP4 & 8k & 8 & 331.6 & 912 & 2414 & 3665 \\
    Qwen3-8B & 2x & NVFP4 & 32k & 8 & 73.6 & 28696 & 29879 & 30172 \\
    Gemma3-12B & 2x & NVFP4 & 8k & 8 & 282.7 & 2907 & 4188 & 4918 \\
    Gemma3-12B & 2x & NVFP4 & 32k & 4 & 36.3 & 11254 & 14412 & 15528 \\
    Gemma3-12B & 2x & NVFP4 & 64k & 4 & 37.9 & 29736 & 31651 & 39082 \\
    Gemma3-27B & 2x & W4A16 & 8k & 4 & 57.9 & 17648 & 26376 & 26439 \\
    Gemma3-27B & 2x & NVFP4 & 8k & -- & X & X & X & X \\
    \midrule
    \multicolumn{9}{l}{\textit{RTX 5060 Ti}} \\
    Qwen3-8B & 1x & NVFP4 & 8k & 8 & 115.3 & 9658 & 16533 & 22776 \\
    Qwen3-8B & 1x & NVFP4 & 16k & 4 & 51.8 & 12365 & 20535 & 23517 \\
    Gemma3-12B & 1x & W4A16 & 8k & -- & X & X & X & X \\
    Gemma3-12B & 1x & NVFP4 & 8k & -- & X & X & X & X \\
    Qwen3-8B & 2x & NVFP4 & 8k & 8 & 158.5 & 2641 & 6315 & 8047 \\
    Qwen3-8B & 2x & NVFP4 & 16k & 4 & 72.4 & 4674 & 6620 & 8710 \\
    Qwen3-8B & 2x & NVFP4 & 32k & 4 & 35.5 & 10329 & 12449 & 22257 \\
    Gemma3-12B & 2x & W4A16 & 8k & 8 & 97.1 & 7443 & 13888 & 23639 \\
    Gemma3-12B & 2x & NVFP4 & 8k & 8 & 135.1 & 7148 & 9747 & 14696 \\
    Gemma3-12B & 2x & NVFP4 & 16k & 4 & 72.4 & 3083 & 5095 & 6058 \\
    Gemma3-27B & 2x & W4A16 & 8k & 4 & 39.4 & 29361 & 46264 & 46726 \\
    Gemma3-27B & 2x & NVFP4 & 8k & 4 & 24.0 & 67207 & 69496 & 71275 \\
    GPT-OSS-20B & 2x & MXFP4 & 8k & 8 & 145.3 & 1570 & 2892 & 3002 \\
  \bottomrule
\end{tabularx}
\end{table}

\begin{table}[H]
\centering\footnotesize
\caption{API workload results (256 tokens input/output, high concurrency).}
\label{tab:api-throughput}
\begin{tabularx}{\textwidth}{X l l c c c c c}
    \toprule
    Model & GPU Config & Precision & Conc. & TPS & TTFT & P95 & P99 \\
    \midrule
    Qwen3-8B & RTX 5090 1x & NVFP4 & 32 & 2676.6 & 77 & 198 & 210 \\
    Qwen3-8B & RTX 5090 1x & NVFP4 & 64 & 4677.7 & 131 & 344 & 402 \\
    Qwen3-8B & RTX 5090 1x & NVFP4 & 128 & 6894.2 & 177 & 606 & 613 \\
    Qwen3-8B & RTX 5090 2x & NVFP4 & 64 & 4466.0 & 173 & 480 & 704 \\
    Qwen3-8B & RTX 5090 2x & NVFP4 & 128 & 6409.8 & 284 & 911 & 1063 \\
    Qwen3-8B & RTX 5090 2x & NVFP4 & 256 & 7438.1 & 599 & 1136 & 1138 \\
    Gemma3-12B & RTX 5090 1x & NVFP4 & 32 & 1570.6 & 113 & 269 & 271 \\
    Gemma3-12B & RTX 5090 1x & NVFP4 & 64 & 2413.4 & 181 & 409 & 537 \\
    Gemma3-27B & RTX 5090 2x & NVFP4 & 32 & 631.1 & 1183 & 5569 & 5573 \\
    GPT-OSS-20B & RTX 5090 1x & MXFP4 & 32 & 3305.0 & 109 & 353 & 357 \\
    GPT-OSS-20B & RTX 5090 1x & MXFP4 & 64 & 4843.9 & 202 & 699 & 708 \\
    \midrule
    Qwen3-8B & RTX 5070 Ti 1x & NVFP4 & 32 & 1986.6 & 170 & 365 & 387 \\
    Qwen3-8B & RTX 5070 Ti 1x & NVFP4 & 64 & 2899.1 & 237 & 620 & 704 \\
    Qwen3-8B & RTX 5070 Ti 1x & NVFP4 & 128 & 3553.6 & 361 & 1067 & 1071 \\
    Qwen3-8B & RTX 5070 Ti 2x & NVFP4 & 64 & 3171.5 & 222 & 683 & 1064 \\
    Qwen3-8B & RTX 5070 Ti 2x & NVFP4 & 128 & X & X & X & X \\
    Gemma3-12B & RTX 5070 Ti 1x & NVFP4 & 32 & 486.4 & 9345 & 11672 & 11764 \\
    Gemma3-12B & RTX 5070 Ti 2x & NVFP4 & 64 & 1049.2 & 2273 & 6063 & 8551 \\
    \midrule
    Qwen3-8B & RTX 5060 Ti 1x & NVFP4 & 32 & 1248.9 & 208 & 687 & 690 \\
    Qwen3-8B & RTX 5060 Ti 1x & NVFP4 & 64 & 1797.5 & 351 & 1025 & 1295 \\
    Qwen3-8B & RTX 5060 Ti 1x & NVFP4 & 128 & 2113.9 & 620 & 1970 & 1974 \\
    Qwen3-8B & RTX 5060 Ti 2x & NVFP4 & 64 & 1646.4 & 556 & 2918 & 2924 \\
    Qwen3-8B & RTX 5060 Ti 2x & NVFP4 & 128 & 2682.4 & 476 & 13552 & 13624 \\
    Gemma3-12B & RTX 5060 Ti 1x & W4A16 & 64 & 305.7 & 38004 & 49834 & 50724 \\
    GPT-OSS-20B & RTX 5060 Ti 1x & MXFP4 & 32 & 487.8 & 7646 & 9540 & 10836 \\
    GPT-OSS-20B & RTX 5060 Ti 1x & MXFP4 & 64 & 488.3 & 22445 & 25697 & 26484 \\
  \bottomrule
\end{tabularx}
\end{table}

\begin{table}[H]
\centering\footnotesize
\caption{Agentic multi-LoRA workload results (Qwen3-8B NVFP4, 3 adapters, 2k context).}
\label{tab:agentic-throughput}
\begin{tabularx}{\textwidth}{X l c c c c c}
    \toprule
    GPU Config & Precision & Conc. & TPS & TTFT & P95 & P99 \\
    \midrule
    RTX 5090 1x & NVFP4 & 16 & 888.8 & 115 & 250 & 724 \\
    RTX 5090 1x & NVFP4 & 32 & 1303.7 & 197 & 645 & 1615 \\
    RTX 5090 1x & NVFP4 & 64 & 1682.7 & 412 & 2340 & 3542 \\
    RTX 5090 2x & NVFP4 & 32 & 1176 & 672 & 4787 & 10848 \\
    RTX 5090 2x & NVFP4 & 64 & 1492 & 1579 & 13794 & 20520 \\
    \midrule
    RTX 5070 Ti 1x & NVFP4 & 8 & 442.2 & 191 & 335 & 888 \\
    RTX 5070 Ti 1x & NVFP4 & 16 & 661.1 & 227 & 582 & 1398 \\
    RTX 5070 Ti 1x & NVFP4 & 32 & 677.5 & 7004 & 10479 & 14940 \\
    RTX 5070 Ti 1x & NVFP4 & 64 & 1044.3 & 11634 & 15029 & 17067 \\
    RTX 5070 Ti 2x & NVFP4 & 32 & 833.4 & 604 & 2367 & 6838 \\
    RTX 5070 Ti 2x & NVFP4 & 64 & 911.0 & 7935 & 21792 & 32518 \\
    \midrule
    RTX 5060 Ti 1x & NVFP4 & 8 & 263.7 & 319 & 478 & 1615 \\
    RTX 5060 Ti 1x & NVFP4 & 16 & 366.1 & 423 & 1067 & 2637 \\
    RTX 5060 Ti 1x & NVFP4 & 32 & 409.1 & 9868 & 17789 & 19621 \\
    RTX 5060 Ti 1x & NVFP4 & 64 & 410.6 & 37077 & 46927 & 50963 \\
    RTX 5060 Ti 2x & NVFP4 & 32 & 506.2 & 998 & 37075 & 39895 \\
    RTX 5060 Ti 2x & NVFP4 & 64 & 546.3 & 2502 & 71943 & 76268 \\
  \bottomrule
\end{tabularx}
\end{table}

\begin{table}[H]
\centering\footnotesize
\caption{Energy efficiency (\$0.12/kWh). Wh/MTok = watt-hours per million tokens.}
\label{tab:energy-efficiency}
\begin{tabularx}{\textwidth}{X l l l c c c c}
    \toprule
    Model & GPU Config & Precision & Workload & Conc. & Wh/MTok & \$/MTok & Power (W) \\
    \midrule
    GPT-OSS-20B & RTX 5090 1x & MXFP4 & RAG-8k & 4 & 266 & 0.032 & 306 \\
    GPT-OSS-20B & RTX 5090 1x & MXFP4 & API & 32 & 42 & 0.005 & 495 \\
    GPT-OSS-20B & RTX 5090 1x & MXFP4 & API & 64 & 25 & 0.003 & 435 \\
    Gemma3-12B & RTX 5090 1x & NVFP4 & API & 32 & 55 & 0.007 & 312 \\
    Gemma3-12B & RTX 5090 1x & NVFP4 & RAG-8k & 8 & 394 & 0.047 & 290 \\
    Gemma3-12B & RTX 5090 1x & W4A16 & RAG-8k & 4 & 459 & 0.055 & 319 \\
    Gemma3-27B & RTX 5090 1x & NVFP4 & RAG-8k & 4 & 864 & 0.104 & 306 \\
    Qwen3-8B & RTX 5090 1x & NVFP4 & RAG-8k & 8 & 239 & 0.029 & 353 \\
    Qwen3-8B & RTX 5090 1x & BF16 & RAG-8k & 8 & 403 & 0.048 & 378 \\
    \midrule
    Gemma3-12B & RTX 5090 2x & NVFP4 & RAG-8k & 8 & 411 & 0.049 & 582 \\
    Gemma3-12B & RTX 5090 2x & W4A16 & RAG-8k & 8 & 550 & 0.066 & 690 \\
    Gemma3-27B & RTX 5090 2x & NVFP4 & RAG-8k & 8 & 716 & 0.086 & 660 \\
    Gemma3-27B & RTX 5090 2x & W4A16 & RAG-8k & 4 & 1284 & 0.154 & 736 \\
    Gemma3-27B & RTX 5090 2x & NVFP4 & API & 32 & 164 & 0.020 & 373 \\
    Qwen3-8B & RTX 5090 2x & NVFP4 & RAG-8k & 8 & 330 & 0.040 & 629 \\
    Qwen3-8B & RTX 5090 2x & NVFP4 & RAG-32k & 8 & 1531 & 0.184 & 811 \\
    Qwen3-8B & RTX 5090 2x & NVFP4 & Agentic & 32 & 100 & 0.012 & 423 \\
    Qwen3-8B & RTX 5090 2x & NVFP4 & Agentic & 64 & 82 & 0.010 & 440 \\
    \midrule
    Gemma3-12B & RTX 5070 Ti 1x & NVFP4 & API & 32 & 114 & 0.014 & 199 \\
    Qwen3-8B & RTX 5070 Ti 1x & NVFP4 & API & 64 & 19 & 0.002 & 203 \\
    Qwen3-8B & RTX 5070 Ti 1x & NVFP4 & API & 128 & 17 & 0.002 & 200 \\
    Qwen3-8B & RTX 5070 Ti 1x & NVFP4 & RAG-8k & 8 & 275 & 0.033 & 208 \\
    Qwen3-8B & RTX 5070 Ti 1x & NVFP4 & RAG-16k & 4 & 605 & 0.073 & 211 \\
    Qwen3-8B & RTX 5070 Ti 1x & NVFP4 & RAG-32k & 4 & 1666 & 0.200 & 204 \\
    Qwen3-8B & RTX 5070 Ti 1x & NVFP4 & Agentic & 32 & 87 & 0.010 & 209 \\
    Qwen3-8B & RTX 5070 Ti 1x & NVFP4 & Agentic & 64 & 69 & 0.008 & 260 \\
    \midrule
    Gemma3-12B & RTX 5070 Ti 2x & NVFP4 & RAG-8k & 8 & 347 & 0.042 & 353 \\
    Gemma3-12B & RTX 5070 Ti 2x & W4A16 & RAG-8k & 8 & 498 & 0.060 & 358 \\
    Qwen3-8B & RTX 5070 Ti 2x & NVFP4 & RAG-8k & 8 & 303 & 0.036 & 361 \\
    Qwen3-8B & RTX 5070 Ti 2x & NVFP4 & RAG-32k & 8 & 1454 & 0.174 & 385 \\
    Gemma3-12B & RTX 5070 Ti 2x & NVFP4 & API & 64 & 72 & 0.009 & 274 \\
    Qwen3-8B & RTX 5070 Ti 2x & NVFP4 & API & 128 & X & X & X \\
    Gemma3-12B & RTX 5070 Ti 2x & NVFP4 & RAG-32k & 4 & 1977 & 0.237 & 258 \\
    Qwen3-8B & RTX 5070 Ti 2x & NVFP4 & Agentic & 64 & 84 & 0.010 & 275 \\
    \midrule
    Gemma3-12B & RTX 5060 Ti 1x & W4A16 & API & 32 & 122 & 0.015 & 135 \\
    Gemma3-12B & RTX 5060 Ti 1x & NVFP4 & API & 32 & 115 & 0.014 & 134 \\
    Qwen3-8B & RTX 5060 Ti 1x & NVFP4 & Agentic & 16 & 105 & 0.013 & 138 \\
    Qwen3-8B & RTX 5060 Ti 1x & NVFP4 & RAG-8k & 8 & 298 & 0.036 & 124 \\
    \midrule
    Gemma3-12B & RTX 5060 Ti 2x & NVFP4 & RAG-8k & 8 & 401 & 0.048 & 195 \\
    Gemma3-12B & RTX 5060 Ti 2x & W4A16 & RAG-8k & 8 & 620 & 0.074 & 216 \\
    Gemma3-27B & RTX 5060 Ti 2x & W4A16 & RAG-8k & 4 & 1530 & 0.184 & 216 \\
    Qwen3-8B & RTX 5060 Ti 2x & NVFP4 & RAG-8k & 8 & 342 & 0.041 & 195 \\
    Qwen3-8B & RTX 5060 Ti 2x & NVFP4 & RAG-16k & 4 & 875 & 0.105 & 228 \\
    Gemma3-12B & RTX 5060 Ti 2x & NVFP4 & RAG-16k & 4 & 885 & 0.106 & 231 \\
    Qwen3-8B & RTX 5060 Ti 2x & NVFP4 & API & 128 & 21 & 0.003 & 200 \\
  \bottomrule
\end{tabularx}
\end{table}

\clearpage
\bibliographystyle{unsrtnat}
\bibliography{refs}

@misc{anthropic2024claude,
  title        = {The {Claude} 3 Model Family: A New Standard for Intelligence},
  author       = {{Anthropic}},
  howpublished = {\url{https://www.anthropic.com/news/claude-3-family}},
  year         = {2024},
  note         = {Accessed: 2024-03-04}
}

@misc{anthropic2025pricing,
  title        = {{Anthropic} {Claude} {API} Pricing},
  author       = {{Anthropic}},
  howpublished = {\url{https://www.anthropic.com/pricing}},
  year         = {2026},
  note         = {Accessed: 2026-01-10. Claude Opus 4.5: \$5/MTok input, \$25/MTok output}
}

@article{bommasani2022foundation,
  title   = {On the Opportunities and Risks of Foundation Models},
  author  = {Rishi Bommasani and Drew A. Hudson and Ehsan Adeli and Russ Altman and Simran Arora and Sydney von Arx and Michael S. Bernstein and Jeannette Bohg and Antoine Bosselut and Emma Brunskill and others},
  journal = {arXiv preprint arXiv:2108.07258},
  year    = {2022}
}

@article{cobbe2021gsm8k,
  title   = {Training Verifiers to Solve Math Word Problems},
  author  = {Karl Cobbe and Vineet Kosaraju and Mohammad Bavarian and Jacob Hilton and Reiichiro Nakano and Christopher Hesse and John Schulman},
  journal = {arXiv preprint arXiv:2110.14168},
  year    = {2021}
}

@article{dettmers2023qlora,
  title   = {{QLoRA}: Efficient Finetuning of Quantized {LLMs}},
  author  = {Tim Dettmers and Artidoro Pagnoni and Ari Holtzman and Luke Zettlemoyer},
  journal = {arXiv preprint arXiv:2305.14314},
  year    = {2023}
}

@misc{eurostat2025ai,
  title        = {Usage of {AI} technologies increasing in {EU} enterprises},
  author       = {{Eurostat}},
  howpublished = {\url{https://ec.europa.eu/eurostat/web/products-eurostat-news/w/ddn-20250123-3}},
  year         = {2025},
  note         = {13.5\% of EU enterprises used AI in 2024 (up from 8\% in 2023); EU Digital Decade target is 75\% by 2030}
}

@misc{eval-harness,
  author    = {Gao, Leo and Tow, Jonathan and Abbasi, Baber and Biderman, Stella and Black, Sid and DiPofi, Anthony and Foster, Charles and Golding, Laurence and Hsu, Jeffrey and Le Noac'h, Alain and Li, Haonan and McDonell, Kyle and Muennighoff, Niklas and Ociepa, Chris and Phang, Jason and Reynolds, Laria and Schoelkopf, Hailey and Skowron, Aviya and Sutawika, Lintang and Tang, Eric and Thite, Anish and Wang, Ben and Wang, Kevin and Zou, Andy},
  title     = {A framework for few-shot language model evaluation},
  month     = 12,
  year      = 2023,
  publisher = {Zenodo},
  version   = {v0.4.0},
  doi       = {10.5281/zenodo.10256836},
  url       = {https://zenodo.org/records/10256836}
}

@misc{gemma32025,
  title        = {Gemma 3 Technical Report},
  author       = {{Gemma Team, Google DeepMind}},
  howpublished = {\url{https://ai.google.dev/gemma}},
  year         = {2025},
  note         = {Released March 12, 2025}
}

@article{google2024gemini,
  title   = {Gemini: A Family of Highly Capable Multimodal Models},
  author  = {{Gemini Team, Google}},
  journal = {arXiv preprint arXiv:2312.11805},
  year    = {2024}
}

@inproceedings{hendrycks2021mmlu,
  title     = {Measuring Massive Multitask Language Understanding},
  author    = {Dan Hendrycks and Collin Burns and Steven Basart and Andy Zou and Mantas Mazeika and Dawn Song and Jacob Steinhardt},
  booktitle = {Proceedings of the International Conference on Learning Representations (ICLR)},
  year      = {2021}
}

@article{hooper2024kvquant,
  title   = {{KVQuant}: Towards 10 Million Context Length {LLM} Inference with {KV} Cache Quantization},
  author  = {Coleman Hooper and Sehoon Kim and Hiva Mohammadzadeh and Michael W. Mahoney and Yakun Sophia Shao and Kurt Keutzer and Amir Gholami},
  journal = {arXiv preprint arXiv:2401.18079},
  year    = {2024}
}

@inproceedings{kwon2023vllm,
  title         = {Efficient Memory Management for Large Language Model Serving with {PagedAttention}},
  author        = {Woosuk Kwon and Shiyi Cao and Siqi Tian and Zhenyu Zhang and Yuandong Tian and Beidi Chen and Tri Dao and Ce Zhang and Ion Stoica},
  booktitle     = {Proceedings of the {ACM} Symposium on Cloud Computing},
  year          = {2023},
  archiveprefix = {arXiv},
  eprint        = {2309.06180}
}

@article{lin2023awq,
  title   = {{AWQ}: Activation-Aware Weight Quantization for {LLM} Compression and Acceleration},
  author  = {Ji Lin and Jiaming Tang and Haotian Tang and Shang Yang and Wei-Ming Chen and Wei-Chen Wang and Guangxuan Xiao and Xingyu Dang and Chuang Gan and Song Han},
  journal = {arXiv preprint arXiv:2306.00978},
  year    = {2023}
}

@misc{lowprecision2025nvfp4,
  title        = {{NVIDIA} Blackwell: The Impact of {NVFP4} For {LLM} Inference},
  author       = {{Edge AI and Vision Alliance}},
  howpublished = {\url{https://www.edge-ai-vision.com/2025/10/nvidia-blackwell-the-impact-of-nvfp4-for-llm-inference/}},
  year         = {2025},
  note         = {Detailed analysis of NVFP4 performance on Blackwell architecture}
}

@misc{marie2025nvfp4,
  title        = {{NVFP4}: Same Accuracy with 2.3x Higher Throughput for 4-Bit {LLMs}},
  author       = {Benjamin Marie},
  howpublished = {\url{https://kaitchup.substack.com/p/nvfp4-same-accuracy-with-23-higher}},
  year         = {2024},
  note         = {Accessed: 2024-11-18}
}

@inproceedings{mattson2020mlperf,
  title     = {{MLPerf} Training Benchmark},
  author    = {Peter Mattson and Christine Cheng and Gregory Diamos and Cody Coleman and Paulius Micikevicius and David Patterson and Hanlin Tang and Gu-Yeon Wei and Peter Bailis and Victor Bittorf and David Brooks and Dehao Chen and Debo Dutta and Udit Gupta and Kim Hazelwood and Andrew Hock and Xinyuan Huang and Daniel Kang and David Kanter and Naveen Kumar and Jeffery Liao and Deepak Narayanan and Tayo Oguntebi and Gennady Pekhimenko and Lillian Pentecost and Vijay Janapa Reddi and Taylor Robie and Tom St. John and Carole-Jean Wu and Lingjie Xu and Cliff Young and Matei Zaharia},
  booktitle = {Proceedings of Machine Learning and Systems (MLSys)},
  year      = {2020}
}

@misc{microsoft2024mxfp4,
  title        = {{MXFP4}: Microscaling Data Formats for Deep Learning},
  author       = {{Microsoft Corporation and Meta and AMD and Intel and NVIDIA and ARM and Qualcomm}},
  howpublished = {\url{https://www.opencompute.org/documents/ocp-microscaling-formats-mx-v1-0-spec-final-pdf}},
  year         = {2024},
  note         = {OCP Microscaling Formats (MX) Specification}
}

@misc{neuralmagic2024fp8,
  title        = {{vLLM} Brings {FP8} Inference to the Open Source Community},
  author       = {{Neural Magic and Red Hat}},
  howpublished = {\url{https://developers.redhat.com/articles/2024/07/15/vllm-brings-fp8-inference-open-source-community}},
  year         = {2024},
  note         = {Accuracy evaluation shows $>$99\% preservation on Open LLM Leaderboard v1 tasks (ARC-c, HellaSwag, MMLU, TruthfulQA, WinoGrande, GSM8k)}
}

@misc{news2025gptoss_tomshardware,
  title        = {OpenAI Intros Two Open-Weight Language Models That Can Run on Consumer {GPUs}},
  author       = {{Tom's Hardware}},
  howpublished = {\url{https://www.tomshardware.com/tech-industry/artificial-intelligence/openai-intros-two-lightweight-open-model-language-models-that-can-run-on-consumer-gpus-optimized-to-run-on-devices-with-just-16gb-of-memory}},
  year         = {2025},
  note         = {Accessed: 2025-11-18}
}

@misc{nvidia_dcgm_docs,
  title        = {{NVIDIA} Data Center {GPU} Manager ({DCGM}) Documentation},
  author       = {{NVIDIA Corporation}},
  howpublished = {\url{https://docs.nvidia.com/datacenter/dcgm/}},
  year         = {2025},
  note         = {Accessed: 2025-11-18}
}

@misc{nvidia_dcgm_exporter,
  title        = {dcgm-exporter},
  author       = {{NVIDIA Corporation}},
  howpublished = {\url{https://github.com/NVIDIA/dcgm-exporter}},
  year         = {2025},
  note         = {Accessed: 2025-11-18}
}

@misc{nvidia_dcgm_exporter_docs,
  title        = {DCGM-Exporter: GPU Telemetry with Prometheus},
  author       = {{NVIDIA Corporation}},
  howpublished = {\url{https://docs.nvidia.com/datacenter/dcgm/latest/gpu-telemetry/dcgm-exporter.html}},
  year         = {2025},
  note         = {Accessed: 2025-11-18}
}

@misc{nvidia2024aiperf,
  title        = {{NVIDIA} {NIM} {LLM} Benchmarking with {AIPerf}},
  author       = {{NVIDIA Corporation}},
  howpublished = {\url{https://docs.nvidia.com/nim/benchmarking/llm/latest/overview.html}},
  year         = {2024},
  note         = {Official NVIDIA documentation for LLM inference benchmarking}
}

@misc{nvidia2025nvfp4,
  title        = {Introducing {NVFP4} for Efficient and Accurate Low-Precision Inference},
  author       = {Eduardo Alvarez and Omri Almog and Eric Chung and Simon Layton and Dusan Stosic and Ronny Krashinsky and Kyle Aubrey},
  howpublished = {\url{https://developer.nvidia.com/blog/introducing-nvfp4-for-efficient-and-accurate-low-precision-inference/}},
  year         = {2024},
  note         = {NVIDIA Developer Blog, Accessed: 2024-11-18}
}

@misc{nvidia5090product,
  title        = {GeForce {RTX} 5090 Graphics Cards},
  author       = {{NVIDIA Corporation}},
  howpublished = {\url{https://www.nvidia.com/en-us/geforce/graphics-cards/50-series/rtx-5090/}},
  year         = {2025},
  note         = {Accessed: 2025-11-18}
}

@article{openai2023gpt4,
  title   = {{GPT-4} Technical Report},
  author  = {{OpenAI}},
  journal = {arXiv preprint arXiv:2303.08774},
  year    = {2023}
}

@misc{openai2025gptoss,
  title        = {Introducing gpt-oss},
  author       = {{OpenAI}},
  howpublished = {\url{https://openai.com/index/introducing-gpt-oss/}},
  year         = {2025},
  note         = {Released August 5, 2025}
}

@misc{openai2025gptossmodelcard,
  title        = {gpt-oss-120b \& gpt-oss-20b Model Card},
  author       = {{OpenAI}},
  howpublished = {\url{https://openai.com/index/gpt-oss-model-card/}},
  year         = {2025},
  note         = {Released August 5, 2025}
}

@misc{openai2025pricing,
  title        = {{OpenAI} {API} Pricing},
  author       = {{OpenAI}},
  howpublished = {\url{https://platform.openai.com/docs/pricing}},
  year         = {2026},
  note         = {Accessed: 2026-01-10. GPT-5 nano: \$0.05/MTok input, \$0.40/MTok output; GPT-5 mini: \$0.25/MTok input, \$2.00/MTok output}
}

@misc{google2025pricing,
  title        = {{Google} {AI} for Developers Pricing},
  author       = {{Google}},
  howpublished = {\url{https://ai.google.dev/pricing}},
  year         = {2026},
  note         = {Accessed: 2026-01-10. Gemini 2.0 Flash-Lite: \$0.075/MTok input, \$0.30/MTok output}
}

@article{peng2023yarn,
  title   = {{YaRN}: Efficient Context Window Extension of Large Language Models},
  author  = {Bowen Peng and Jeffrey Quesnelle and Honglu Fan and Enrico Shippole},
  journal = {arXiv preprint arXiv:2309.00071},
  year    = {2023}
}

@article{shoeybi2019megatron,
  title   = {Megatron-{LM}: Training Multi-Billion Parameter Language Models Using Model Parallelism},
  author  = {Mohammad Shoeybi and Mostofa Patwary and Raul Puri and Patrick LeGresley and Jared Casper and Bryan Catanzaro},
  journal = {arXiv preprint arXiv:1909.08053},
  year    = {2019}
}

@misc{yang2025qwen3,
  title        = {{Qwen3} Technical Report},
  author       = {An Yang and Alibaba Cloud Team},
  howpublished = {\url{https://qwenlm.github.io/}},
  year         = {2025},
  note         = {Released April 28, 2025}
}

@inproceedings{zellers2019hellaswag,
  title     = {{HellaSwag}: Can a Machine Really Finish Your Sentence?},
  author    = {Rowan Zellers and Ari Holtzman and Yonatan Bisk and Ali Farhadi and Yejin Choi},
  booktitle = {Proceedings of the 57th Annual Meeting of the Association for Computational Linguistics},
  year      = {2019}
}

\end{document}